%% file: main.tex
\documentclass[10pt,twocolumn,letterpaper]{article}

\usepackage{cvpr}              

\usepackage{graphicx}
\usepackage{amsmath}
\usepackage{amssymb}
\usepackage{booktabs}

\usepackage{nicefrac}
\usepackage{paralist}
\usepackage{tabularx}
\usepackage{multirow, multicol}
\usepackage{booktabs}
\usepackage{makecell}
\usepackage[dvipsnames]{xcolor}
\usepackage{xcolor, colortbl}

\usepackage{soul}

\usepackage[accsupp]{axessibility}  

\include{macro}

\usepackage[pagebackref,breaklinks,colorlinks, citecolor=blue]{hyperref}

\renewcommand{\paragraph}[1]{\noindent \textbf{#1}}
\usepackage[capitalize]{cleveref}
\crefname{section}{Sec.}{Secs.}
\Crefname{section}{Section}{Sections}
\Crefname{table}{Table}{Tables}
\crefname{table}{Tab.}{Tabs.}

\newcommand{\figref}[1]{Fig.~\ref{#1}}
\newcommand{\tabref}[1]{Table~\ref{#1}}
\newcommand{\secref}[1]{Section~\ref{#1}}

\def\cvprPaperID{5690} 
\def\confName{CVPR}
\def\confYear{2022}

\begin{document}

\title{AP-BSN: Self-Supervised Denoising for Real-World Images \protect\\ via Asymmetric PD and Blind-Spot Network}
\author{
Wooseok Lee ~~~~~~~~
Sanghyun Son ~~~~~~~~
Kyoung Mu Lee\\
ASRI, Department of ECE, Seoul National University\\
{\tt\small adntjr4@gmail.com, \{thstkdgus35, kyoungmu\}@snu.ac.kr}
}
\maketitle

\begin{abstract}
    Blind-spot network (BSN) and its variants have made significant advances in self-supervised denoising.
    Nevertheless, they are still bound to synthetic noisy inputs due to less practical assumptions like pixel-wise independent noise.
    Hence, it is challenging to deal with spatially correlated real-world noise using self-supervised BSN.
    Recently, pixel-shuffle downsampling (PD) has been proposed to remove the spatial correlation of real-world noise.
    However, it is not trivial to integrate PD and BSN directly, which prevents the fully self-supervised denoising model on real-world images.
    We propose an Asymmetric PD (AP) to address this issue, which introduces different PD stride factors for training and inference.
    We systematically demonstrate that the proposed AP can resolve inherent trade-offs caused by specific PD stride factors and make BSN applicable to practical scenarios.
    To this end, we develop AP-BSN, a state-of-the-art self-supervised denoising method for real-world sRGB images. 
    We further propose random-replacing refinement, which significantly improves the performance of our AP-BSN without any additional parameters.
    Extensive studies demonstrate that our method outperforms the other self-supervised and even unpaired denoising methods by a large margin, without using any additional knowledge, e.g., noise level, regarding the underlying unknown noise.
    \urlstyle{same}
    \blfootnote{Code is available at: \url{https://github.com/wooseoklee4/AP-BSN}}
\end{abstract}

\input{main/1}

\input{main/2}

\input{main/3}

\input{main/4}
\input{main/5}

\input{main/6}

\section*{Acknowledgement}
This work was supported in part by IITP grant funded by the Korea government (MSIT) [No. 2021-0-01343, Artificial Intelligence Graduate School Program (Seoul National University), and No.2021-0-02068, Artificial Intelligence Innovation Hub]

\renewcommand{\thetable}{S\arabic{table}}
\renewcommand{\thefigure}{S\arabic{figure}}
\renewcommand{\theequation}{S\arabic{equation}}
\renewcommand{\thesection}{S\arabic{section}}

\setcounter{table}{0}
\setcounter{figure}{0}
\setcounter{equation}{0}
\setcounter{section}{0}

\twocolumn[{
\includegraphics[width=\linewidth]{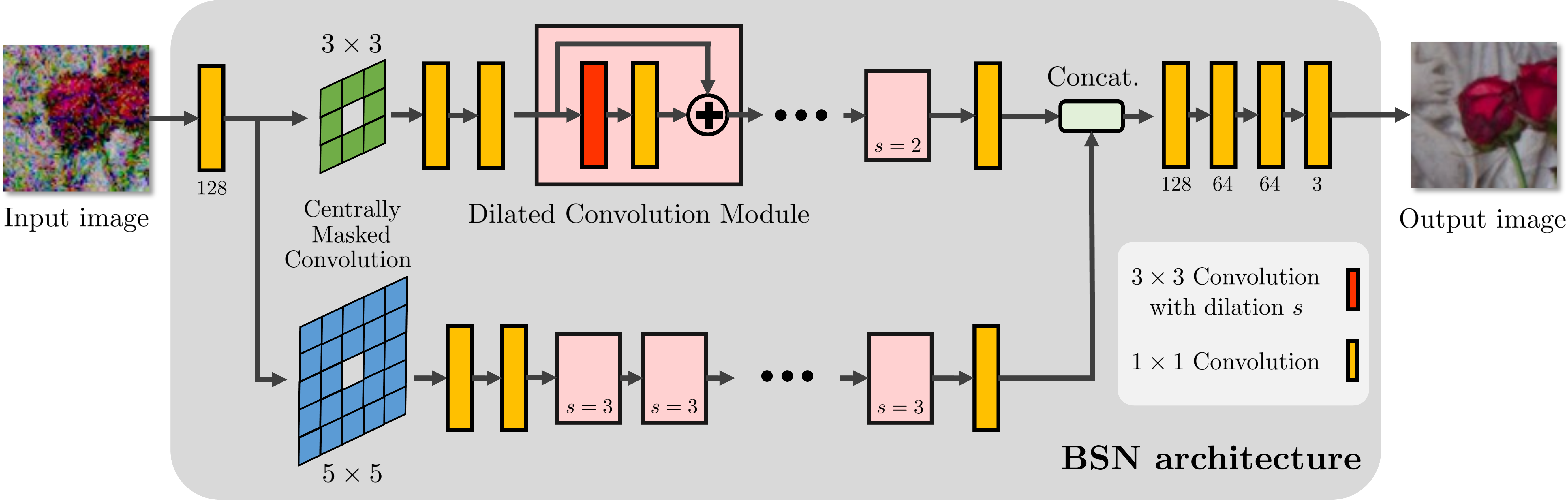}
\captionof{figure}{
    \textbf{Visualization of our BSN architecture.}
    We adopt $3 \times 3$ and $5 \times 5$ Centrally Masked Convolutions~\cite{DBSN} to implement the blind-spot network.
    Each Dilated Convolution module (DC) contains one $3 \times 3$ dilated convolution with a stride $s$, where $s = 2$ and $s = 3$ are used for the upper and lower path of the network, respectively.
    For each path, we stack 9 DC modules.
    The number of output channels is denoted below each convolutional layer, where 128 is used by default.
} \label{fig:network}
\vspace{4mm}
}]

\input{supple/implementation}


\input{supple/ablation}

\input{supple/nind}

\input{supple/more}

{\small
\bibliographystyle{ieee_fullname}
\bibliography{egbib}
}

\end{document}

%% file: macro.tex
\newcommand{\img}[1]{\text{I}_{\:\!\text{#1}}}
\newcommand{\imge}[1]{\text{I}_{#1}}
\newcommand{\ap}[2]{\text{AP}_{#1\text{/}#2}}
\newcommand{\pd}[1]{\text{PD}_#1}

\newcommand{\bsn}{B}
\newcommand{\rrpp}{\text{R}^\text{3}}

\newcommand{\normtwo}[1]{\left\lVert #1 \right\rVert^2_2}
\newcommand{\normone}[1]{\left\lVert #1 \right\rVert_1}
\newcommand{\paren}[1]{\left( #1 \right)}

\newcommand{\code}[1]{{\fontfamily{qcr}\selectfont #1}}

\newcommand{\proposed}{AP-BSN}

\newcommand\blfootnote[1]{%
    \begingroup%
\let\thefootnote\relax\footnotetext{\hspace{-4pt}#1}%
\endgroup}%

%% file: main/1.tex
\section{Introduction}
\label{sec:intro}

\input{main/figs/fig_first}

Image denoising is one of the essential topics in the computer vision area, which aims to recover a clean image from the noisy signal.
Due to its practical usage in several vision-related applications, several learning-based denoising algorithms~\cite{DnCNN, FFDNet, RED30, MemNet} have been proposed with the advent of convolutional neural networks (CNNs).
Conventional methods usually adopt additive white Gaussian noise (AWGN) to acquire large-scale training data by synthesizing clean-noisy image pairs for supervised learning.
Nevertheless, models learned on the synthetic noise do not generalize well in practice since the characteristics of real-world noise differ much from AWGN.
To overcome the limitation, several attempts have been made to construct pairs of real-world datasets like SIDD~\cite{SIDD} and NIND~\cite{NIND}.
Using the real-world training pairs, supervised denoising methods~\cite{VDN, AINDNet, DANet, P3AN, NBNet} can be trained to restore clean images from the noisy real-world input.
However, constructing the real-world dataset requires massive human labor, strictly controlled environments, and complicated post-processing.
In addition, it is difficult to generalize the learned model toward diverse practical scenarios as the characteristic of noise varies much for the different capturing devices.

Recently, several self-supervised approaches~\cite{Noise2Void, Noise2Self, Noise2Same, Laine2019, Neighbor2Neighbor, R2R, NAC} have been introduced, which do not rely on paired training data.
Such methods require noisy images only for training instead of clean-noisy pairs.
Among them, Blind-Spot Network~(BSN)~\cite{Noise2Void} is one of the representative methods motivated by Noise2Noise~\cite{Noise2Noise}.
Under the assumption that noise signals are pixel-wise independent and zero-mean, BSN reconstructs a clean pixel from the neighboring noisy pixels without referring to the corresponding input pixel.
Based on BSN, several approaches~\cite{Laine2019, DBSN, honzatko2020efficient, krull2020probabilistic} have achieved better performance on synthetic noise while ensuring strict blindness \wrt the center pixel.
However, real-world noises are known to be spatially-correlated~\cite{park2009case, jin2020review, chatterjee2011noise}, which does not meet the basic assumption of BSN: noise is pixel-wise independent.

To break spatial correlation of real-world noise, Zhou~\etal~\cite{whenAWGN} utilize pixel-shuffle downsampling (PD).
PD creates a mosaic by subsampling a noisy image with a fixed stride factor, and thereby increases an actual distance between noise signals.
Nevertheless, integrating PD to BSN is nontrivial when handling real-world noise in a fully self-supervised manner, where it cannot stand alone without knowledge from additional noisy-clean synthetic pairs~\cite{DBSN}.
We identify that the principal reason for such limitation is the trade-off between the pixel-wise independent assumption and reconstruction quality.
For example, a large PD stride factor ($ > 3$) ensures the strict pixel-wise independent noise assumption and benefits BSN during training.
However, it also destructs detailed structures and textures from the noisy image.
In contrast, a small PD stride factor ($ \leq 3$) preserves image structures but cannot satisfy the pixel-wise independent assumption when training BSN.

Inspired by these observations, we propose Asymmetric PD (\textbf{AP}), which uses different stride factors for training and inference.
For real-world noise, we systematically validate that a specific combination of training and inference strides can compensate shortcomings of each other.
Then, we integrate AP to BSN (\textbf{AP-BSN}), which can learn to denoise noisy real-world inputs in a fully self-supervised manner, without requiring any prior knowledge of underlying noise.
Furthermore, we propose random-replacing refinement (\textbf{$\rrpp$}), a novel post-processing method that improves the performance of our AP-BSN without any additional training.
To the best of our knowledge, our \proposed{} is the first attempt to introduce self-supervised BSN for real-world sRGB noisy images.
Extensive studies demonstrate that our method outperforms not only the state-of-the-art self-supervised denoising methods but also several unsupervised/unpaired approaches by a large margin.
We summarize our contributions as follows:
\begin{compactitem}[$\bullet$]
    \item To handle spatially correlated real-world noise in a blind fashion, we propose a novel self-supervised \textbf{AP-BSN}.
    Our framework employs asymmetric PD stride factors for training and inference in conjunction with BSN.
    \item We propose random-replacing refinement ($\textbf{R}^3$), a novel post-processing method that further improves our AP-BSN without any additional parameters.
    \item Our \proposed{} is the first self-supervised BSN that covers real-world sRGB noisy inputs and outperforms the other self-supervised and even several unpaired solutions by large margins.
\end{compactitem}

%% file: main/figs/fig_first.tex
\begin{figure}[t]
    \renewcommand{\wp}{0.495\linewidth}
    \captionsetup[subfloat]{font=small, justification=centering}
    
    \centering

    \subfloat[DnCNN~\cite{DnCNN} \\ \small{( Supervised )}]
    {\includegraphics[width=\wp]{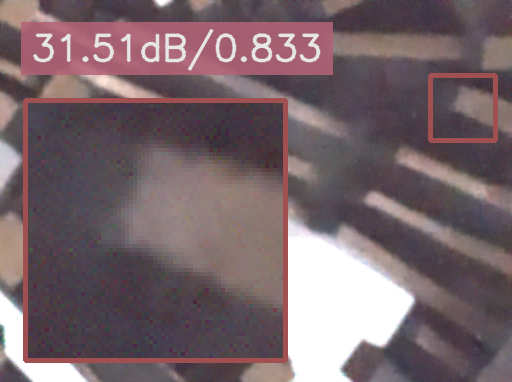}\label{fig:first_fig_DnCNN}}
    \hfill
    \subfloat[C2N~\cite{C2N} $+$ DIDN~\cite{DIDN} \\ \small{( Unpaired )}]
    {\includegraphics[width=\wp]{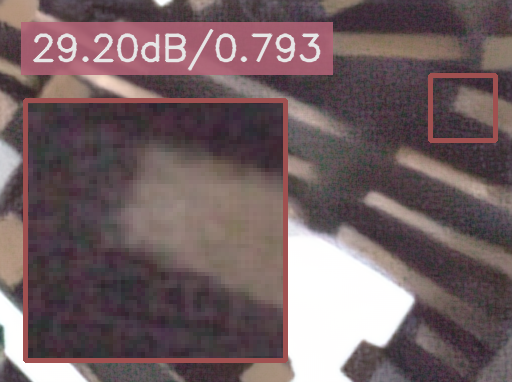}\label{fig:first_fig_C2N}}
    \\
    \subfloat[NAC~\cite{NAC} \\ \small{( Self-supervised )}]
    {\includegraphics[width=\wp]{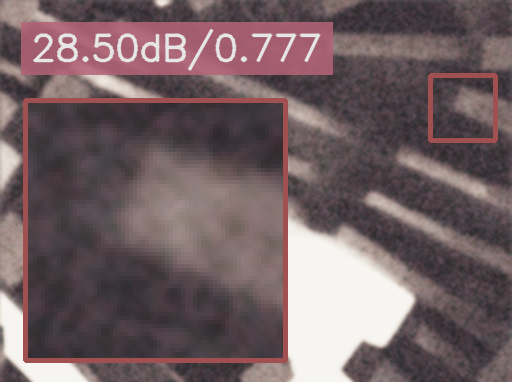}\label{fig:first_fig_NAC}}
    \hfill
    \subfloat[\textbf{AP-BSN $+$ $\rrpp$ (Ours)}  \\ \small{( Self-supervised )}]
    {\includegraphics[width=\wp]{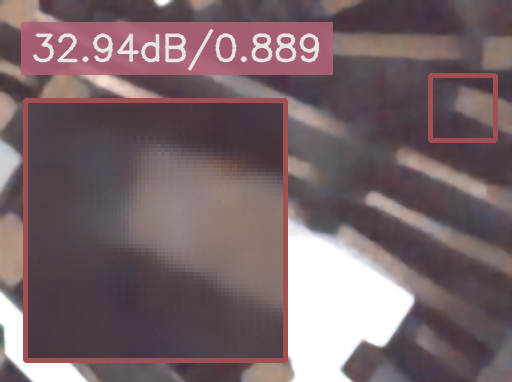}\label{fig:first_fig_ours}}
    \\
    \vspace{-2mm}
    \caption{
        \textbf{Visual comparison between different denoising methods on the DND benchmark~\cite{DND}.}
        (a) DnCNN is trained on real-world noisy-clean pairs from the SIDD~\cite{SIDD} dataset.
        (b) C2N uses clean SIDD~\cite{SIDD} and noisy DND~\cite{DND} samples to simulate real-world noise distribution in an unsupervised manner.
        (c--d) Self-supervised methods can be trained on the DND~\cite{DND} noisy images directly.
        We mark PSNR(dB) and SSIM with respect to the ground-truth clean image for the quantitative comparison.
    }
    \vspace{-6mm}
    \label{fig:first_fig}
\end{figure}

%% file: main/2.tex
\section{Related Work}
\label{sec:rel_works}

\paragraph{Deep image denoising for synthetic noise.}
    Beyond the classical non-learning based approaches~\cite{BM3D, WNNM, K-SVD, EPLL}, DnCNN \cite{DnCNN} has introduced a CNN-based architecture to remove AWGN from a given image.
    Following DnCNN, several learning-based approaches have been proposed such as FFDNet~\cite{FFDNet}, RED30~\cite{RED30}, and MemNet~\cite{MemNet}, with advanced network architectures.
    Nevertheless, the methods trained on AWGN suffer from generalization toward the real-world denoising due to domain discrepancy between real and synthetic noises.
    Specifically, Guo~\etal~\cite{CBDNet} have demonstrated that AWGN-based denoisers do not perform well when input noise signals are signal-dependent~\cite{poigau} or spatially-correlated~\cite{park2009case, chatterjee2011noise, jin2020review}.

\paragraph{Real-world image denoising.}
    To reduce the gap between synthetic and real-world denoising, CBDNet~\cite{CBDNet} simulates in-camera ISP with gamma correction and demosaicking process.
    Then, synthetic heteroscedastic Gaussian noise can be transformed into realistic noise signals, which can be used to generate training pairs for supervised learning.
    Zhou~\etal~\cite{whenAWGN} have proposed pixel-shuffle downsampling (PD) to cover spatially-correlated real-world noise with conventional AWGN denoisers.
    In contrast, there have been a few attempts to capture the noisy-clean training pairs from real-world~\cite{SIDD, NIND}.
    Using the real-world pairs, it is straightforward to train supervised denoising methods~\cite{VDN, AINDNet, DANet, P3AN, NBNet}, which generalize well on the corresponding real-world inputs.
    However, constructing real-world pairs require huge labor and is not always available.

\paragraph{Unpaired image denoising.}
    When sets of unpaired clean and real-world noisy images are available, several methods leverage generative approaches~\cite{gans} to synthesize realistic noise from the clean samples~\cite{GCBD, GAN2GAN, UIDNet, C2N}.
    Among them, GCBD~\cite{GCBD} selectively uses plain regions from noisy images for stable learning.
    Recently, C2N~\cite{C2N} explicitly considers various noise characteristics to simulate real-world noise more accurately.
    Using the generated noisy-clean pairs, the following supervised denoising model~\cite{DnCNN, DIDN} can be trained to deal with real-world noise.
    On the other hand, Wu~\etal~\cite{DBSN} distill knowledge from a self-supervised denoising model while adopting synthetic noisy-clean pairs.
    Still, it is important to match the scene statistics of clean and noisy datasets even in the unpaired configuration~\cite{C2N}, which can be difficult in practice.

\paragraph{Self-supervised denoising.}
    A major bottleneck for real-world denoising is the absence of appropriate training data.
    Therefore, several approaches have been proposed to train their model using noisy images \emph{only}.
    Motivated by Noise2Noise~\cite{Noise2Noise}, Noise2Void~\cite{Noise2Void} and Noise2Self~\cite{Noise2Self} have introduced novel self-supervised learning frameworks by masking a portion of noisy pixels from the input image.
    Notably, the concept of BSN~\cite{Noise2Void} has been later extended to more efficient architectures in the form of four halved receptive fields~\cite{Laine2019} or dilated and masked convolutions~\cite{DBSN}.
    While Noise2Same~\cite{Noise2Same} does not use BSN, a novel loss term is used to satisfy $\mathcal{J}$-invariant property~\cite{Noise2Self} in the denoising network.
    Neighbor2Neighbor~\cite{Neighbor2Neighbor}, on the other hand, acquires the noisy-noisy pair for self-supervision by subsampling the given input.
    Nevertheless, the above self-supervised methods heavily rely on assumptions that noise signals are pixel-wise independent.
    Therefore, they usually end up learning \emph{identity} mappings when applied to real-world sRGB images as noise signals are spatially-correlated~\cite{park2009case, chatterjee2011noise, jin2020review}.

    Recent Noisier2Noise~\cite{Noisier2Noise}, NAC~\cite{NAC}, and R2R~\cite{R2R} add different synthetic noise signals to the given input to make auxiliary training pairs.
    However, Noisier2Noise requires prior knowledge regarding the underlying noise distribution, and Noisy-As-Clean relies on weak noise assumptions.
    R2R also requires several prior information such as noise level and ISP function, which may not be available in real-world scenarios.

%% file: main/3.tex
\section{BSN and PD}
\label{sec:bsnpd}

\paragraph{Blind-spot network.}
BSN~\cite{Noise2Void} is a variant of the conventional CNN that does not \emph{see} the center pixel in the receptive field to predict the corresponding output pixel.
Several studies~\cite{Noise2Void, Noise2Self, krull2020probabilistic} have demonstrated that BSN $B \paren{ \cdot }$ can learn to denoise a noisy image $\img{N} \in \mathbb{R}^{H \times W} $ in a self-supervised manner.
We note that the image has a resolution of $ H \times W $, and color channels are omitted for simplicity.
To train BSN, the following two assumptions must be satisfied: noise is spatially, \ie, pixel-wise, independent and zero-mean.
Under such assumptions, it is known~\cite{Noise2Same, Noise2Self} that minimizing the self-supervised loss $\mathcal{L}_\text{self}$ \wrt BSN is equivalent to conventional supervised learning as follows:
\begin{equation}
    \begin{split}
        \mathcal{L}_\text{self} &= \mathbb{E}_{\img{N}} \normtwo{ \bsn \paren{ \img{N} } - \img{N} } \\
         &= \mathbb{E}_{ \img{N}, \img{C} } \normtwo{ \bsn \paren{ \img{N} }  - \img{C} } + c = \mathcal{L}_\text{super} + c,
    \end{split}
    \label{eq:bsneq}
\end{equation}
where $\img{C} \in \mathbb{R}^{H \times W} $ is a clean ground-truth for the noisy input $\img{N}$, $\mathcal{L}_\text{super}$ is a supervised denoising loss function, and $c$ is a constant, respectively.

Therefore, several types of BSN~\cite{Laine2019, DBSN} are constructed under the pixel-wise independent noise assumption.
However, real-world noise is spatially correlated due to the image signal processors (ISP).
Specifically, demosaicking on Bayer filter~\cite{park2009case, chatterjee2011noise, jin2020review} involves interpolation between noisy subpixels.
\figref{fig:noise_char} demonstrates that in real-world, noise intensities between neighboring pixels show non-negligible correlation based on their relative distance.
Since the neighboring noise signals can be clues for inferring the unseen center pixel, we have identified that BSN operates as an approximately identity mapping on real-world sRGB images.

\input{main/figs/fig_noise_char}

\paragraph{Pixel-shuffle downsampling.}
Zhou~\etal~\cite{whenAWGN} have introduced a novel concept of PD to break down the spatial correlation in the real-world noise.
Specifically, $\text{PD}_s$ can be regarded as an inverse operation of the pixel-shuffling~\cite{shi2016real} with a stride factor of $s$.
Since real-world noise signals are correlated with few neighboring pixels, subsampling in PD process may break the dependency between them.
Then, conventional denoising algorithms can be applied to the downsampled images, where the PD-inverse operation $\text{PD}_s^{-1}$ follows to reconstruct a full-sized output.
To preserve image textures and details, Zhou~\etal~\cite{whenAWGN} set the stride factor to 2, \ie $\pd{2}$, for the best performance.

\input{main/figs/fig_problem}

%% file: main/figs/fig_noise_char.tex
\begin{figure}[t!]
    \renewcommand{\wp}{0.49\linewidth}
    \renewcommand{\vs}{-2mm}
    \definecolor{camred}{RGB}{214, 39, 40}
    \definecolor{camviolet}{RGB}{148, 103, 189}
    \captionsetup[subfloat]{font=normalsize}
    \centering
    \subfloat[By relative distances $d$]
    {\includegraphics[width=\wp]{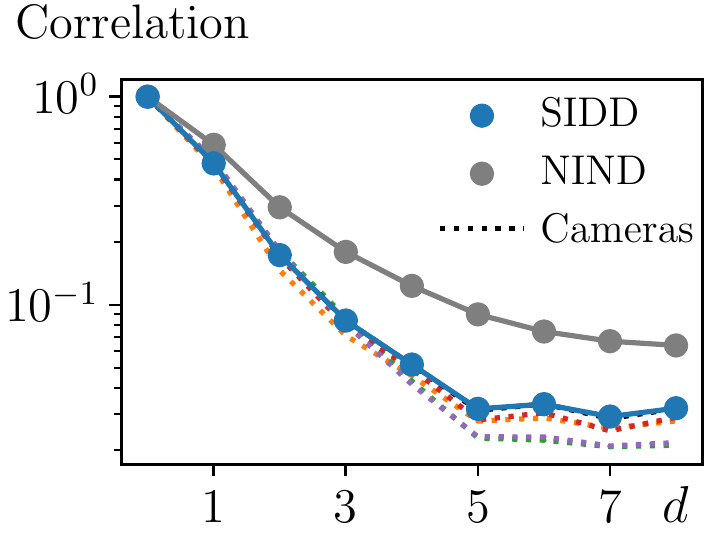}\label{fig:noise_char_b}}
    \hfill
    \subfloat[By relative locations]
    {\includegraphics[width=\wp]{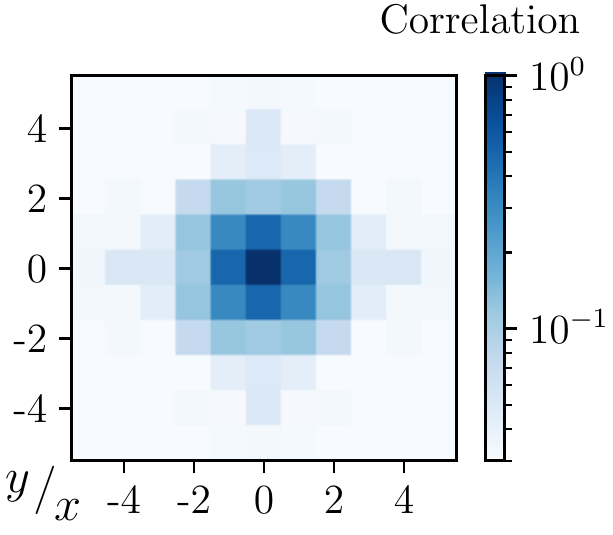}\label{fig:noise_char_a}}
    \\
    \vspace{-2mm}
    \caption{
        \textbf{Analysis of spatial correlation on real-world noise.}
        (a) As the relative distance $d$ between two noise signals increases, their correlation decreases.
        We note that different camera devices, \eg, \textcolor{camred}{\textbf{Motorola Nexus 6 (N6)}} or \textcolor{camviolet}{\textbf{LG G4}}, in the SIDD~\cite{SIDD} dataset show similar noise behaviors in terms of spatial correlation, as illustrated with dotted lines.
        (b) $x$ and $y$ axis represent a relative distance along with horizontal and vertical directions, respectively.
    }
    \label{fig:noise_char}
    \vspace{-4mm}
\end{figure}

%% file: main/figs/fig_problem.tex
\begin{figure}[t!]
    \renewcommand{\wp}{0.495\linewidth}
    \captionsetup[subfloat]{font=small}
    \centering
    \subfloat[Real-world noisy image $\img{N}$]
    {\includegraphics[width=\wp]{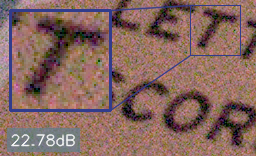}\label{fig:prb_pdbsn_real}}
    \hfill
    \subfloat[Clean image $\img{C}$]
    {\includegraphics[width=\wp]{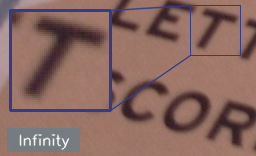}\label{fig:prb_pdbsn_clean}}
    \\
    \subfloat[$\text{PD}_2$-BSN]
    {\includegraphics[width=\wp]{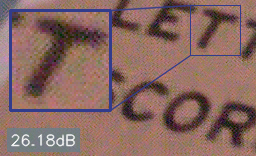}\label{fig:prb_pdbsn_bsnpd2}}
    \hfill
    \subfloat[$\text{PD}_5$-BSN]
    {\includegraphics[width=\wp]{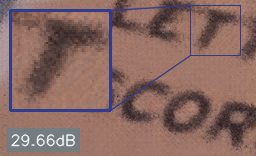}\label{fig:prb_pdbsn_bsnpd5}}
    \\
    \subfloat[Zhou~\etal~\cite{whenAWGN}]{\includegraphics[width=\wp]{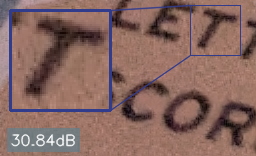}\label{fig:prb_pdbsn_pd}}
    \hfill
    \subfloat[\textbf{\proposed{} + $\rrpp$ (Ours)}] {\includegraphics[width=\wp]{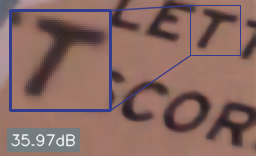}\label{fig:prb_pdbsn_ours}}
    \\
    \vspace{-2mm}
    \caption{
        \textbf{Issues on $\text{PD}_s$-BSN when handling real-world noise.}
        (c) With a small stride factor, PD-BSN cannot remove noise from the input $\img{N}$.
        (d) With a large stride factor, PD-BSN destructs edge structures.
        (e) When AWGN denoiser meets PD~\cite{whenAWGN}, the model cannot completely remove real-world noise.
        (f) Our self-supervised approach delivers an accurate denoising result by overcoming the limitation of combining PD and BSN.
    }
    \label{fig:prb_pdbsn}
    \vspace{-4mm}
\end{figure}
%

%% file: main/4.tex
\section{Method}
\label{sec:method}
Our goal is to generalize BSN on real-world sRGB images in a self-supervised manner.
To this end, we adopt PD and minimize the following loss $\mathcal{L}_\text{BSN}$ to train BSN:
\begin{equation}
    \begin{split}
        \mathcal{L}_\text{BSN} &= \normone{ \text{PD}_s^{-1} \paren{ B \paren{ \text{PD}_s \paren{ \img{N} } } } - \img{N} } \\
        &= \normone{ \img{BSN}^s - \img{N} },
    \end{split}
    \label{eq:pd_bsn}
\end{equation}
where $\img{BSN}^s$ is an output from $\text{PD}_s$ and BSN pipeline, namely $\textbf{PD}_s$-\textbf{BSN}.
Instead of widely-used $L^2$ loss, we use $L^1$ norm for better generalization~\cite{EDSR}.
In brief, we first decompose the given noisy image $\img{N}$ into $s^2$ sub-images.
We note that $\text{PD}_s \paren{ \img{N} }$ is a tiling of those sub-images~\cite{whenAWGN} $\img{sub}^s \in \mathbb{R}^{\nicefrac{H}{s} \times \nicefrac{W}{s}}$, as shown in \figref{fig:pd}.
Then, we apply BSN to the sub-images and reconstruct the output $\img{BSN}^s$ using the PD-inverse operation $\text{PD}_s^{-1}$.

However, it is not straightforward to apply PD-BSN directly on real-world sRGB images.
While Wu~\etal~\cite{DBSN} have also tried to integrate PD and BSN, they resort to knowledge distillation combined with additional synthetic noisy-clean pairs.
We have also observed that PD-BSN is not applicable to real-world noisy images when trained with the self-supervised loss in Eq.~\eqref{eq:pd_bsn}.
Figs.~\ref{fig:prb_pdbsn_bsnpd2} and \ref{fig:prb_pdbsn_bsnpd5} demonstrate that $\pd{2}$-BSN and $\pd{5}$-BSN cannot restore a clean and sharp image from the given noisy input, regardless of the PD stride factor $s$.

\subsection{Trade-offs in PD-BSN}
\label{ssec:tradeoff}
When applying the AWGN-based denoiser on real-world images, Zhou~\etal~\cite{whenAWGN} use $\pd{2}$.
However, we have observed that PD exhibits different behaviors as the stride factor $s$ varies.
Therefore, we first describe two important aspects of PD-BSN regarding the stride factor $s$.

\paragraph{Breaking spatial correlation.}
Originally, PD has been proposed to reduce spatial correlation between neighboring noise signals in real-world images.
While Zhou~\etal~\cite{whenAWGN} resort to the stride factor of 2, our analysis in \figref{fig:noise_char_b} demonstrates that the stride factor should be at least 5 to minimize the dependency in the given noise signal.
In other words, noise signals in the sub-images $\img{sub}^2$ are still spatially correlated, where the pixel-wise independent noise assumption for BSN does not hold.

\paragraph{Aliasing artifacts.}
Nevertheless, the sub-images $\img{sub}^s$ from $\text{PD}_s$ suffer stronger degree of aliasing as the stride factor $s$ becomes larger.
From the perspective of signal processing, it is well-known that a downsampled image suffers aliasing when the original signal is not properly bandlimited~\cite{oppenheim1997signals}.
Since the PD process does not leverage a low-pass filter before subsampling, we have identified that aliasing occurs as a form of noise when applying large-stride PD, \eg, $s = 5$, as shown in \figref{fig:pd}.

\input{main/figs/fig_pd}

\input{main/figs/fig_main}

\subsection{Effective training stride factor for PD-BSN}
\label{ssec:pd_training}
We next establish a strategy to \emph{train} $\text{PD}_s$-BSN.
For such purpose, the correlation between noise signals in the training input images $\img{N}$ has to be minimized~\cite{Noise2Void}.
However, as discussed in \secref{ssec:tradeoff}, $\pd{2}$ is not enough to break spatial correlation of real-world noise.
Since the underlying assumption of BSN is not satisfied, the model \emph{cannot} learn to denoise with $\pd{2}$.
By setting $s = 5$ to suppress the spatial correlation between noise signals in training samples, we \emph{can} train BSN on the smaller sub-images $\img{sub}^5$.

We note that BSN also learns to remove the aliasing artifacts induced by the large PD stride factor.
The aliasing happens when high-frequency signals are not removed before subsampling~\cite{oppenheim1997signals}.
As the high-frequency components change rapidly in the original noisy image $\img{N}$, we can \emph{ignore} the spatial correlation of aliasing artifacts in the sub-images $\img{sub}^5$.
The artifacts also satisfy the \emph{zero-mean} constraint, \ie, their statistical mean is approximately the same as that of the noisy image $\img{N}$, since they are random samples of the observed signal.
As the aliasing artifacts satisfy two preconditions of BSN, our PD-BSN also learns to remove them.

\subsection{Asymmetric PD for BSN}
Several studies~\cite{GCBD, C2N} have already identified that matching data distribution between training and test samples play a critical role in accurate image denoising.
Therefore, it is natural to use the same stride factor for training and inference when applying PD-BSN.
However, we have found that the learned BSN recognizes aliasing artifacts from $\pd{5}$ as noise signals to be removed during \emph{inference}.
Since those artifacts contain necessary information to reconstruct high-frequency details, $\pd{5}$-BSN destructs image structures during inference while removing noise as shown in \figref{fig:prb_pdbsn_bsnpd5}.

Instead, we propose an asymmetric stride factor during the \emph{inference} of PD-BSN, which we refer to as Asymmetric PD ($\ap{a}{b}$).
We note that $a$ and $b$ are stride factors for training and inference, respectively.
Specifically, we set $b = 2$ so that the sub-images $\img{sub}^2$ contain minimum aliasing artifacts during inference, while the correlation between neighboring noise signals can be decreased.
In \secref{sec:experiment}, we demonstrate how each trade-off, \ie, spatial correlation and aliasing artifacts, affects the denoising performance of our method.
Our BSN with the proposed $\ap{5}{2}$ (\proposed{}) can learn to remove real-world noise in a self-supervised manner, while preserving image structures as shown in \figref{fig:prb_pdbsn_ours}.
We also note that our \proposed{} does not require any clean samples for training and is directly applicable to sRGB noisy images in practical scenarios.
\figref{fig:main_fig} illustrates our asymmetric training and inference schemes for \proposed{}.

\input{main/figs/fig_rrpp}

\subsection{Random-replacing refinement}
Even with the smallest stride factor, PD and the following denoising step may remove some informative high-frequency components from the input, resulting in visual artifacts~\cite{whenAWGN}.
Therefore, Zhou~\etal~\cite{whenAWGN} propose PD-refinement to suppress artifacts from the PD process and enhance details of the denoising result.
In PD-refinement, an $i$-th replaced image $I_{\mathcal{M}_i}$ is formulated as follows:
\begin{equation}
    \imge{\mathcal{M}_i} = \mathcal{M}_i \odot \img{N} + (\mathbf{1}- \mathcal{M}_i) \odot \img{BSN}^s,
    \label{eq:replacemen}
\end{equation}
where $\mathcal{M}_i \in \left\{ 0, 1 \right\}^{H \times W}$ is a binary mask indicating pixels to be replaced and $\odot$ denotes element-wise multiplication.
Here, $\mathcal{M}_i$ is a structured binary matrix where ones are placed with a fixed stride of 2 and $\sum_i \mathcal{M}_i = \mathbf{1}$.
After the replacement, each image $I_{\mathcal{M}_i}$ is denoised again and averaged to reconstruct the final result $\img{DN}$ as follows:
\begin{equation}
    \img{DN} = \frac{1}{T} \sum_{i = 1}^{T}{D \paren{\imge{\mathcal{M}_i}}},
    \label{eq:refinement}
\end{equation}
where $D$ is the denoising model targeting pixel-wise independent noise and $T$ is the number of masks, \ie, $2^2 = 4$, for the original PD-refinement.

However, the deterministic strategy in PD-refinement leaves a non-negligible correlation between the replaced noise signals. 
Specifically, a replaced noisy pixel in $\imge{ \mathcal{M}_i }$ is \emph{always} correlated with some of its neighbors, as visualized in \figref{fig:post_compare_correlation}.
Such correlation negatively affects the performance of the following denoising method $D$, which assumes spatially uncorrelated noise.
Therefore, we propose an advanced \textbf{random-replacing refinement} ($\textbf{R}^3$) strategy to mitigate the limitation of PD-refinement.

In our $\rrpp$, we adopt $T$ randomized binary masks $\mathcal{R}_i$ instead, which are defined as follows:
\begin{equation}
    \mathcal{R}_i \paren{ x, y } =
        \begin{cases}
            1, & \text{with a probability of } p, \\
            0, & \text{otherwise},
        \end{cases}
    \label{eq:rrpp_mask}
\end{equation}
where $\paren{ x, y }$ denotes an index of the element in a $H \times W$ matrix.
For Eq.~\eqref{eq:replacemen} and Eq.~\eqref{eq:refinement}, we adopt the randomized mask $\mathcal{R}_i$ rather than the fixed one $\mathcal{M}_i$ to acquire the final output.
Since noisy pixels are randomly placed in the $i$-th replaced image $\imge{\mathcal{R}_i}$, an expected correlation between two noise signals is multiplied by $p$, as shown in \figref{fig:post_compare_correlation}.
Hence, our $\rrpp{}$ significantly reduces the expected correlation compared to the previous PD-refinement.
When we combine $\rrpp{}$ with AP-BSN, we do not perform PD and feed the replaced image $\imge{\mathcal{R}_i}$ to BSN directly because spatial correlation of noise in the input is almost negligible.
\figref{fig:post_compare} highlights major differences between PD-refinement and our $\rrpp$.

%% file: main/figs/fig_pd.tex
\begin{figure}[t]
    \renewcommand{\wp}{\linewidth}
    \definecolor{myblue}{RGB}{0, 112, 192}
    \centering
    \includegraphics[width=\linewidth]{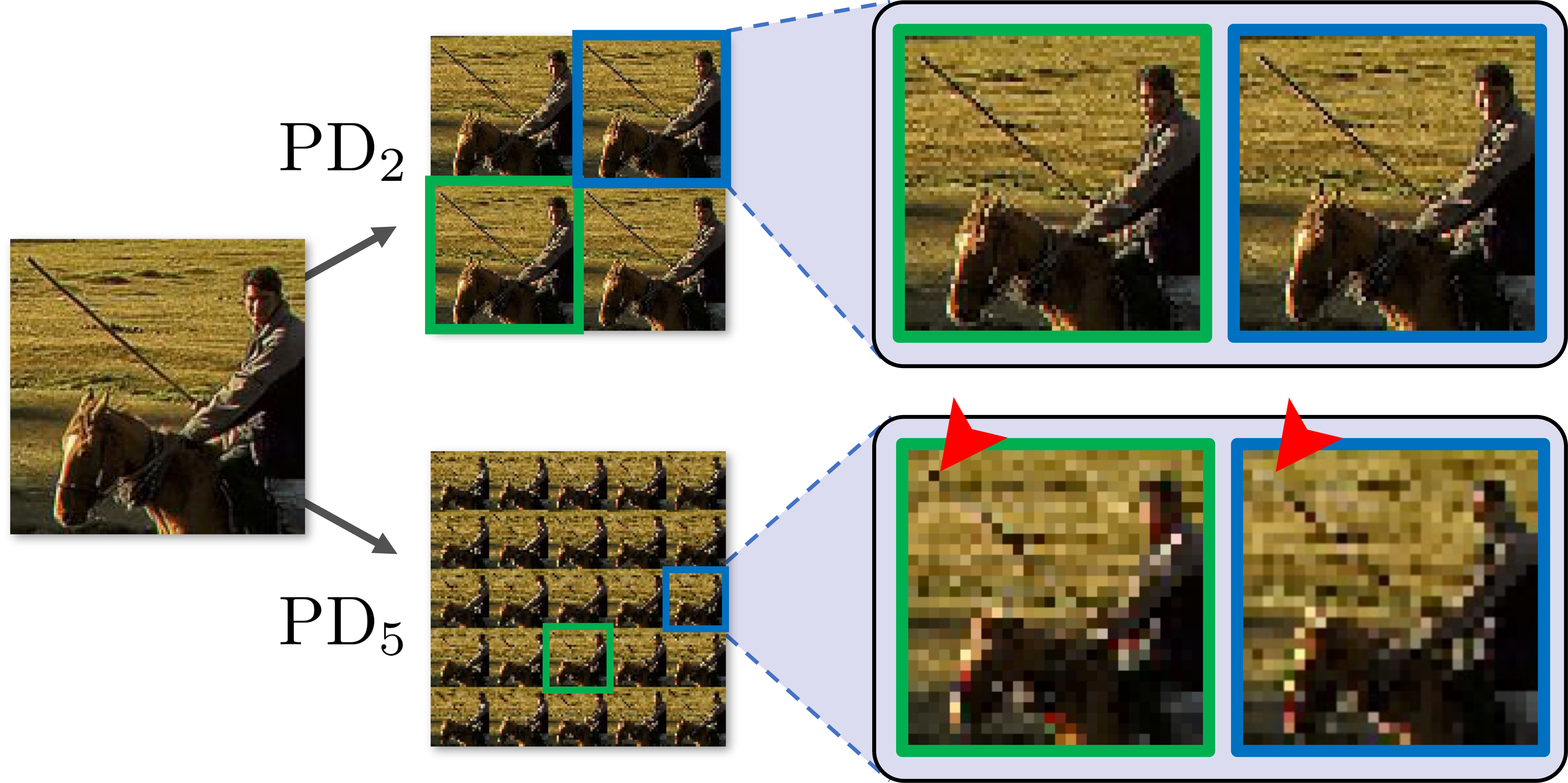}
    \\
    \vspace{-2mm}
    \caption{
        \textbf{Comparison between $\pd{2}$ and $\pd{5}$.}
        Each operation decomposes the given image into 4 and 25 sub-images, respectively.
        In sub-images from $\pd{5}$, we mark the aliasing artifact, \ie a black dot, with \textcolor{red}{\textbf{red}}, which can be interpreted as noise for BSN.
        We note that the artifact does not appear in the \textcolor{myblue}{\textbf{blue}} sub-image.
    }
    \label{fig:pd}
    \vspace{-4mm}
\end{figure}

%% file: main/figs/fig_main.tex
\begin{figure*}[t]
    \renewcommand{\wp}{1.0\linewidth}
    \centering
    \includegraphics[width=\wp]{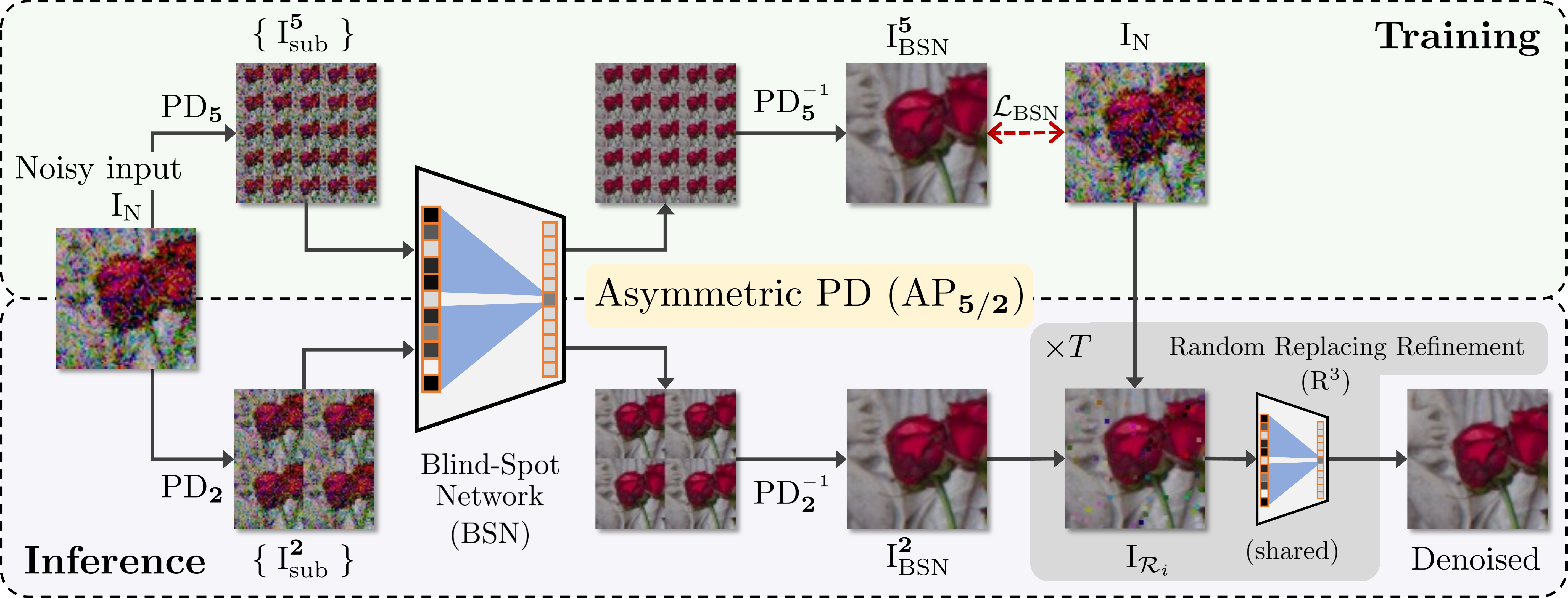}
    \\
    \vspace{-2mm}
    \caption{
        \textbf{Overview of the proposed \proposed{} and $\rrpp$ post-processing.}
        We visualize the proposed $\ap{5}{2}$-BSN.
        To apply BSN on real-world sRGB images, we introduce $\ap{a}{b}$ to maximize synergies of using different stride factors for training and inference.
        %
        %
        %
        We use a large stride factor, \eg, $a = 5$, to ensure pixel-wise independence between noise signals for training.
        During the inference, we use a minimum stride factor of $b = 2$ to avoid aliasing artifacts while breaking down the spatial correlation of noise to some extent.
        %
        %
        %
        Our random-replacing refinement ($\rrpp$) further improves the performance of AP-BSN without any additional parameters.
    }
    \label{fig:main_fig}
    \vspace{-4mm}
\end{figure*}

%% file: main/figs/fig_rrpp.tex
\begin{figure}[t!]
    \renewcommand{\wp}{\linewidth}
    \captionsetup[subfloat]{font=small}
    \vspace{-2mm}
    \centering
    \subfloat[Expected correlations]{\includegraphics[width=0.435\wp]{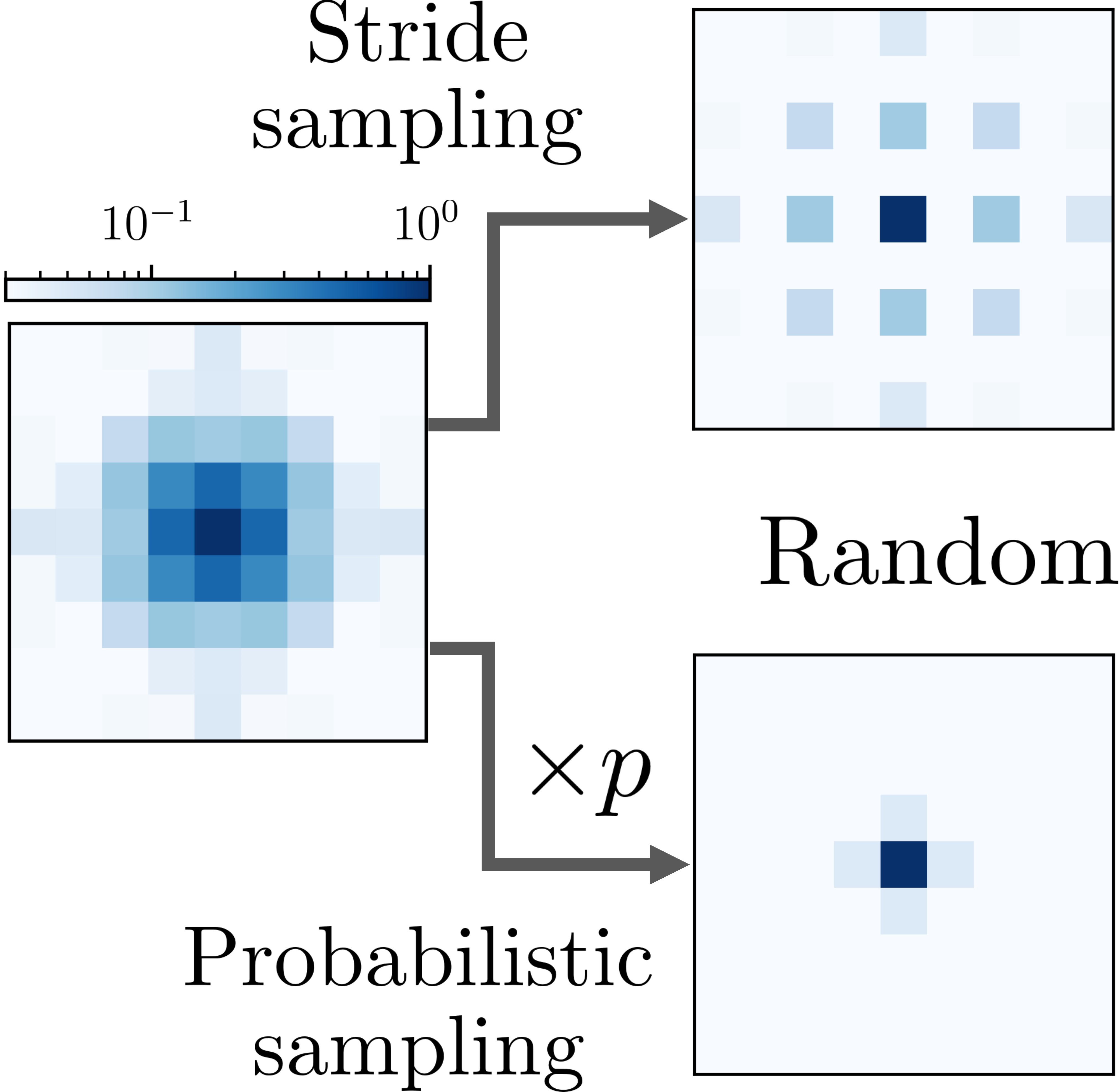}\label{fig:post_compare_correlation}}
    \hfill
    \subfloat[Replacement strategies]{\includegraphics[width=0.555\wp]{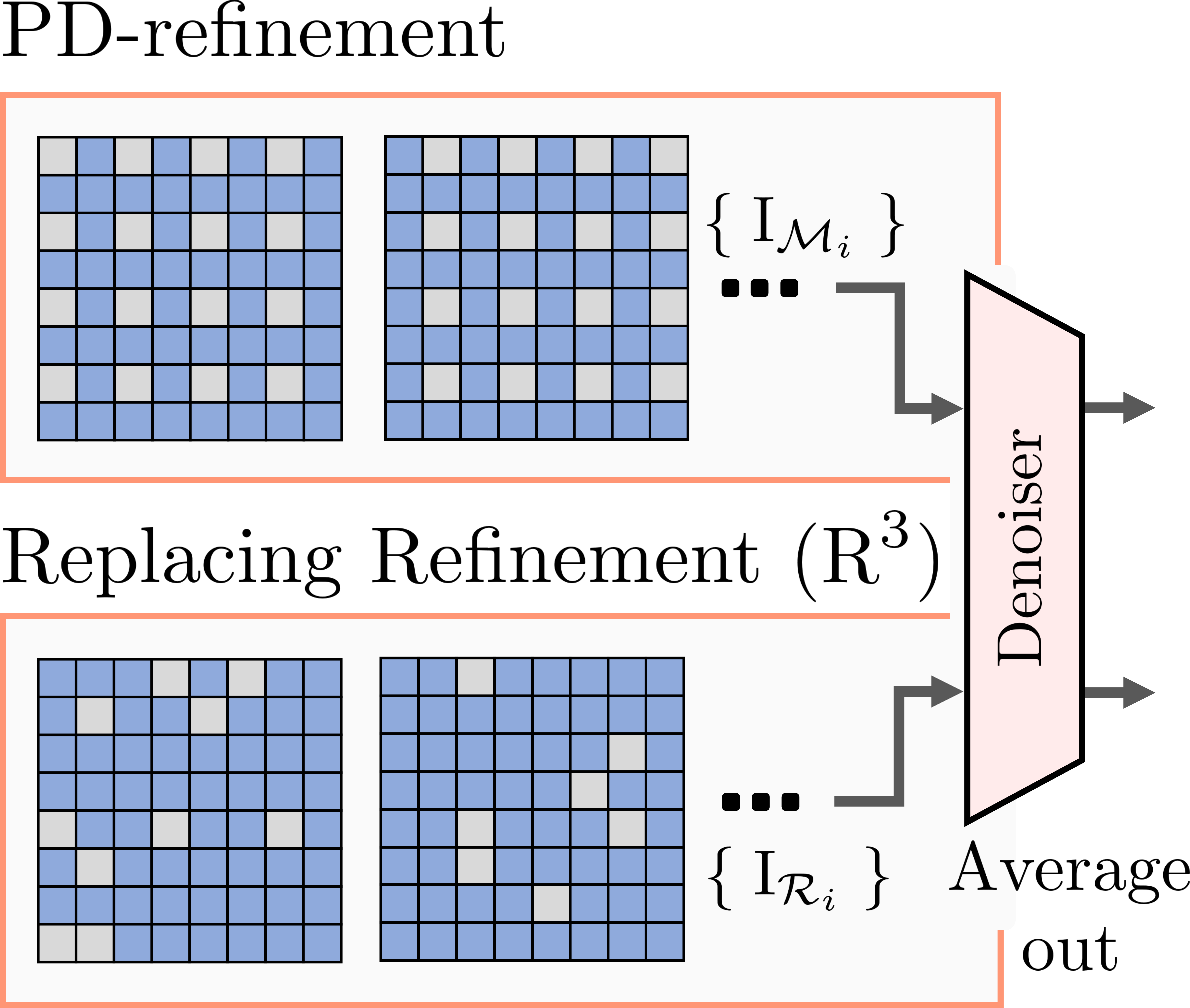}\label{fig:post_compare_method}}
    \\
    \vspace{-2mm}
    \caption{
        \textbf{Comparison between PD-refinement and our $\rrpp$.}
        While PD-refinement adopts regular binary masks $\mathcal{M}_i$ with a stride of $2$, our $\rrpp{}$ uses randomized masks $\mathcal{R}_i$.
        (a) We compare the expected spatial correlation of noise signals in the replaced image $\imge{\mathcal{M}_i}$ and $\imge{\mathcal{R}_i}$.
        (b) Each gray box represents a pixel from the original noisy image $\img{N}$, which replaces the denoised pixel in $\img{BSN}^s$.
        %
        %
        %
        %
    }
    \label{fig:post_compare}
    \vspace{-4mm}
\end{figure}

%% file: main/5.tex
\section{Experiment}
\label{sec:experiment}

\subsection{Experimental configurations}
\paragraph{Dataset.}
To train and evaluate our AP-BSN, we adopt widely-used real-world image denoising datasets: SIDD~\cite{SIDD} and DND~\cite{DND}.
SIDD-Medium consists of 320 real-world noisy and clean image pairs for training.
For validation and performance evaluation, we adopt SIDD validation and benchmark datasets, respectively.
Both contain 1,280 noisy patches with a size of $256 \times 256$, where the corresponding clean images are also provided for the validation set.

The DND dataset does not include training images and consists of 50 real-world noisy inputs only for evaluation.
Rather than using the SIDD-Medium training dataset for this case, we enjoy the advantage of a fully self-supervised learning framework and use the same data for training and performance evaluation.
In other words, we train our AP-BSN on 50 noisy DND images and reconstruct the final denoising results from the same inputs.

\input{main/figs/fig_pdsf_graph}
\input{main/figs/fig_pdsf}

\input{main/figs/tab_main}

\paragraph{Metric.}
To evaluate our \proposed{} and compare it with the other denoising methods, we introduce widely-used peak signal-to-noise ratio (PSNR) and structural similarity (SSIM) metrics.
For SIDD and DND benchmarks, we upload our results to the evaluation sites to calculate the metrics.
On the SIDD validation dataset, we use the corresponding functions in \code{skimage.metrics} library and RGB color space for comparison.

\paragraph{Implementation and optimization.}
We use PyTorch 1.9.0~\cite{fw_pytorch} for implementation.
By default, we adopt $\ap{5}{2}$ and set $p$ and $T$ to 0.16 and 8, respectively, for the proposed $\rrpp$.
For BSN, we modify the architecture from Wu~\etal~\cite{DBSN} for efficiency.
\proposed{} is trained using Adam~\cite{Adam} optimizer, and the initial learning rate starts from $10^{-4}$.
More details are described in our supplementary material.

\subsection{Analyzing Asymmetric PD}
\label{sec:experiment_stride}
We first validate the effect of AP for real-world sRGB denoising.
To this end, we conduct an extensive study on all possible combinations of feasible stride factors, \ie, $a \in \left\{ 2, 3, 4, 5, 6 \right\}$ and $b \in \left\{ 1, 2, 3, 4, 5, 6 \right\}$, in \figref{fig:pdsf_graph_psnr}.
We note that BSN cannot be trained when $a = 2$ due to the spatial correlation of real-world noise.
With larger training stride factors $a$, the input noise of BSN follows pixel-wise independent assumption more strictly.
Therefore, the model can learn the denoising function better, where the performances are maximized with $a = 5$.
When $a = 6$ is used, $\ap{6}{b}$-BSN performs slightly worse since the noise in the SIDD~\cite{SIDD} dataset show increasing correlation as shown in \figref{fig:noise_char_b}.
Interestingly, $a = 6$ is slightly better than $a = 5$ on the NIND~\cite{NIND} dataset, as the correlation gradually decreases \wrt to the relative distance between pixels.
More analysis on the NIND dataset is reported in our supplementary material.
During the inference, BSN cannot remove real-world noise without PD, \ie, $b = 1$, as it is learned on pixel-wise independent noise.
The performances are maximized when $b = 2$, as the trade-off between spatial correlation and aliasing can be optimized.
With larger inference stride factors, \ie, $b > 2$, AP-BSN performs worse because more image details are removed in the form of aliasing artifacts.

In \figref{fig:pdsf_graph_sample}, we justify that the existence of aliasing artifacts is a critical factor for our denoising framework.
When applying $\ap{5}{b}$-BSN to the plain region illustrated in \figref{fig:pdsf_img_plain}, the model performs better as the inference stride factor $b$ becomes larger.
Since the region does not contain high-frequency information, aliasing artifacts do not appear in Figs.~\ref{fig:pdsf_plain51}, \ref{fig:pdsf_plain52}, and \ref{fig:pdsf_plain55}.
Rather, the spatial correlation of noise signals becomes smaller with a larger $b$, which results in better performance.
For a general image in \figref{fig:pdsf_img_textural}, our $\ap{5}{b}$-BSN shows a similar behavior to that of \figref{fig:pdsf_graph_psnr}, while the performance drop is much severe due to the stronger aliasing artifacts as shown in \figref{fig:pdsf_textural55}.

\input{main/figs/fig_rrpp_exp}

\input{main/figs/fig_comparison}

\subsection{Analyzing Random-Replacing Refinement}
\figref{fig:rrpp_exp} shows a detailed ablation study on hyperparameters for the proposed $\rrpp{}$.
We first set $T = 2, 4, 8$ to find the optimal replacement probability $p$.
As shown in \figref{fig:rrpp_exp_p}, our $\rrpp{}$ shows a consistent behavior where the maximum performance is achieved with $p \approx 0.16$.
We note that a larger $p$ increases the expected spatial correlation of noise signals which degrades the performance.
Due to the stochastic behavior, the number of randomized masks $T$ is not limited in our $\rrpp{}$, while PD-refinement can only use $T = 4$.
\figref{fig:rrpp_exp_T} demonstrates that the proposed $\rrpp{}$ performs better than PD-refinement even with $T = 2$, and the performance increases as the number of randomized masks $T$ goes higher.
%
%
Since the computational complexity of $\rrpp{}$ is proportional to $T$, we set $T = 8$ to balance the performance and runtime.

\subsection{AP-BSN for real-world denoising}
Our AP-BSN aims to denoise real-world sRGB images in a self-supervised manner.
\tabref{tab:main} compares various image denoising models on widely-used SIDD and DND benchmark datasets.
Using noisy images $\emph{only}$ for training, the proposed AP-BSN + $\rrpp{}$ achieves the best performance among several unpaired~\cite{C2N, DBSN} and self-supervised approaches.
Especially, we note that self-supervised NAC~\cite{NAC} and R2R~\cite{R2R} are constructed on less practical assumptions like noise level is weak or ISP function is known.
On the other hand, our approach adopts BSN with several observations regarding the properties of PD and real-world noise.
Therefore, we do not rely on specific assumptions and show better generalization on several real-world datasets.
In addition, the proposed $\rrpp{}$ post-processing further improves the evaluation PSNR more than 1dB on the SIDD benchmark track without any additional parameters.
\figref{fig:comparison} provides visual comparisons between several methods addressed in \tabref{tab:main}.

Furthermore, AP-BSN can be trained on noisy samples directly, without using any clean images.
Since several un-/self-supervised methods are trained on auxiliary images~\cite{R2R} or generated noise~\cite{C2N}, the discrepancy between training and test distributions may result in sub-optimal solutions.
In contrast, our approach can use target sRGB noisy images directly during training phase.
To validate the merit of our framework, we train AP-BSN on the SIDD benchmark and evaluate on the \emph{same} dataset.
The last row of \tabref{tab:main} shows that the fully self-supervised strategy improves the denoising performance by about 1dB without making any modifications.
Although SIDD-Medium contains about $\times 60$ more pixels than the benchmark split, such an improvement highlights that AP-BSN can also generalize well on practical cases where there exist noisy test samples only.

%% file: main/figs/fig_pdsf_graph.tex
\begin{figure}[t!]
    \renewcommand{\wp}{\linewidth}
    \renewcommand{\vs}{-2mm}
    \captionsetup[subfloat]{font=small}
    \centering
    \subfloat[Effects of asymmetric $a$/$b$]
    {\includegraphics[width=0.465\wp]{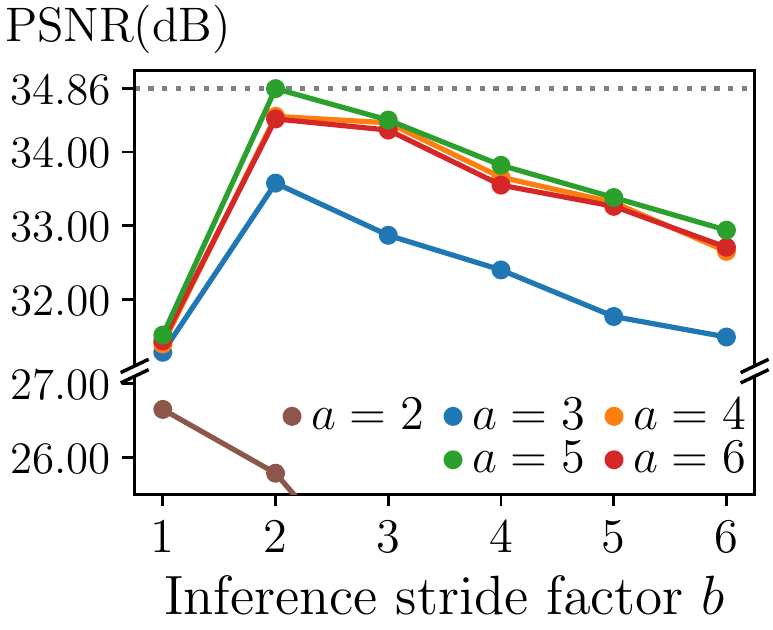}\label{fig:pdsf_graph_psnr}}
    \hfill
    \subfloat[Effects of aliasing artifacts]
    {\includegraphics[width=0.535\wp]{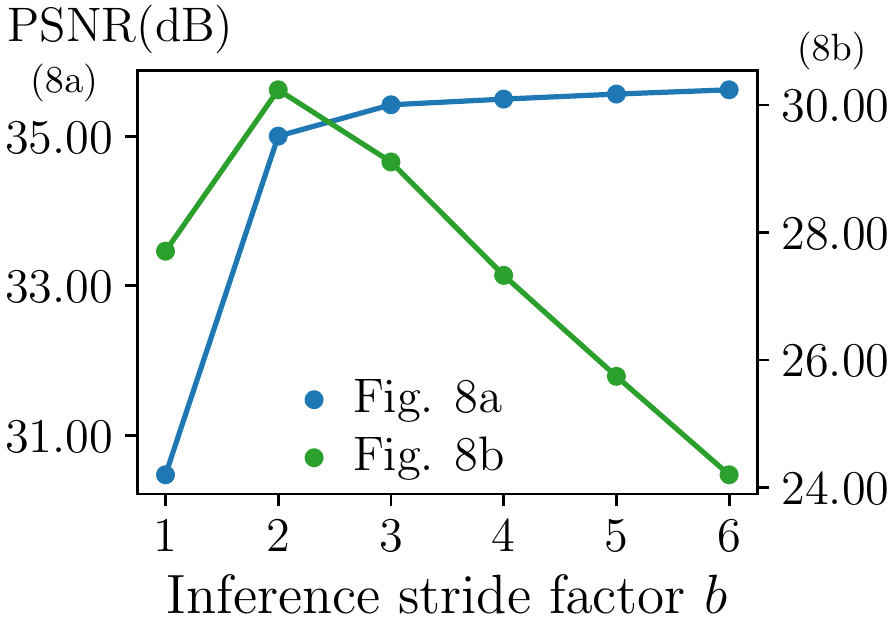}\label{fig:pdsf_graph_sample}}
    \\
    \vspace{-2mm}
    \caption{
    \textbf{Ablation study of $\ap{a}{b}$-BSN on the SIDD validation dataset.}
    We note that the proposed $\rrpp$ post-processing is not applied in these ablation studies.
    (a) Our $\ap{a}{b}$-BSN consistently achieves the best performance when $b = 2$.
    (b) We validate $\ap{5}{b}$-BSN on two representative images displayed in Figs.~\ref{fig:pdsf_img_plain} and \ref{fig:pdsf_img_textural}.
    %
    }
    \label{fig:pdsf_graph}
    \vspace{-2mm}
\end{figure}

%% file: main/figs/fig_pdsf.tex
\begin{figure}[t!]
    \renewcommand{\wp}{0.495\linewidth}
    \newcommand{\wwp}{0.165\linewidth}
    \newcommand{\traininfer}[2]{\footnotesize{Train} \normalsize{$\pd{#1}$}, \footnotesize{Inference} \normalsize{$\pd{#2}$} }
    \captionsetup[subfloat]{font=small}
    \centering
    \subfloat[Plain region]
    {\includegraphics[width=\wp]{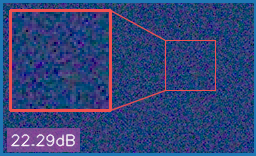}\label{fig:pdsf_img_plain}}
    \hfill
    \subfloat[Textured region]
    {\includegraphics[width=\wp]{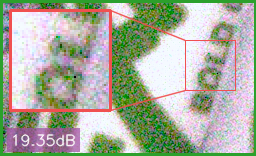}\label{fig:pdsf_img_textural}}
    \\
    \subfloat[$\ap{5}{1}$]
    {\includegraphics[width=\wwp]{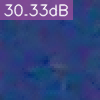}\label{fig:pdsf_plain51}}
    \hfill
    \subfloat[$\ap{5}{2}$]
    {\includegraphics[width=\wwp]{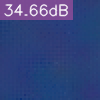}\label{fig:pdsf_plain52}}
    \hfill
    \subfloat[$\ap{5}{5}$]
    {\includegraphics[width=\wwp]{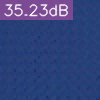}\label{fig:pdsf_plain55}}
    \hfill
    \subfloat[$\ap{5}{1}$]
    {\includegraphics[width=\wwp]{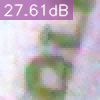}\label{fig:pdsf_textural51}}
    \hfill
    \subfloat[$\ap{5}{2}$]
    {\includegraphics[width=\wwp]{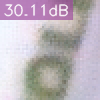}\label{fig:pdsf_textural52}}
    \hfill
    \subfloat[$\ap{5}{5}$]
    {\includegraphics[width=\wwp]{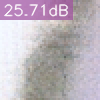}\label{fig:pdsf_textural55}}
    \\
    \vspace{-2mm}
    \caption{
        \textbf{Visual comparison of the trade-off in $\ap{a}{b}$-BSN.}
        %
        %
        (c--e) For a plain region in (a), performance of AP-BSN gradually increases as the inference stride factor $b$ becomes larger.
        (f--h) For a textured region in (b), AP-BSN performs the best when $b = 2$ but shows decreased performance for larger $b$.
        Please refer to \figref{fig:pdsf_graph_sample} for more details.
    }
    \label{fig:pdsf_img}
    \vspace{-4mm}
\end{figure}
%

%% file: main/figs/tab_main.tex
\begin{table*}[t]
    \centering
    \newcommand{\ts}{\,\,}
    \newcommand{\ourspad}{\:}
    \newcommand{\ours}{{^\diamond}}
    \newcommand{\fromRpad}{\:}
    \newcommand{\fromR}{{^\textbf{\tiny{R}}}}
    \newcommand{\selfen}{^*}
    \newcommand{\benchmark}{\dagger}
    
    \small
    
    \begin{tabularx}{\linewidth}{cc >{\raggedright\arraybackslash}X  cccccc}
        \toprule
        & & \multirow{2}{*}{Method} & \multicolumn{2}{c}{SIDD} && \multicolumn{2}{c}{DND} & \\
        & & & PSNR$^\uparrow$(dB) & SSIM$^\uparrow$ && PSNR$^\uparrow$(dB) & SSIM$^\uparrow$  \\
        \midrule
        
        \multirow{2}{*}{\makecell{Non-learning based}} 
        && BM3D~\cite{BM3D} & 25.65 & 0.685 && 34.51 & 0.851 & \\
        && WNNM~\cite{WNNM} & 25.78 & 0.809 && 34.67 & 0.865 & \\
        \midrule
        
        \multirow{3}{*}{\makecell{Supervised \\ (Synthetic pairs)}}
        && DnCNN~\cite{DnCNN} & 23.66 & 0.583 && 32.43 & 0.790 & \\
        && CBDNet~\cite{CBDNet} & 33.28 & 0.868 && 38.05 & 0.942 & \\
        && Zhou~\etal~\cite{whenAWGN} & $\ourspad$ \text{34.00}$\ours$ & $\ourspad$ \text{0.898}$\ours$ && 38.40 & 0.945 & \\
        \midrule
        
        \multirow{4}{*}{\makecell{Supervised \\ (Real pairs)}} 
        && DnCNN~\cite{DnCNN} &  $\ourspad$ \text{35.13}$\ours$ & $\ourspad$ \text{0.896}$\ours$ && $\ourspad$ \text{37.89}$\ours$ & $\ourspad$ \text{0.932}$\ours$ & \\
        && AINDNet (R)$\selfen$~\cite{AINDNet} & 38.84 & 0.951 && 39.34 & 0.952 & \\
        && VDN~\cite{VDN} & 39.26 & 0.955 && 39.38 & 0.952 & \\
        && DANet\cite{DANet} & 39.43 & 0.956 && 39.58 & 0.955 & \\
        \midrule
        
        \multirow{3}{*}{\makecell{Unsupervised \\ (Unpaired)}}
        && GCBD~\cite{GCBD} & - & - && 35.58 & 0.922 & \\
        && C2N~\cite{C2N} $+$ DIDN$\selfen$~\cite{DIDN} & 35.35 & 0.937 && 37.28 & 0.924 & \\
        && D-BSN~\cite{DBSN} $+$ MWCNN~\cite{MWCNN} & - & - && 37.93 & 0.937 & \\
        \midrule
        
        \multirow{7}{*}{\makecell{Self-supervised}}
        && Noise2Void~\cite{Noise2Void} & $\fromRpad$ \text{27.68}$\fromR$ & $\fromRpad$ \text{0.668}$\fromR$ && - & - & \\
        && Noise2Self~\cite{Noise2Self} & $\fromRpad$ \text{29.56}$\fromR$ & $\fromRpad$ \text{0.808}$\fromR$ && - & - & \\
        && NAC~\cite{NAC}    & - & - && 36.20 & 0.925 & \\
        && R2R~\cite{R2R} & 34.78 & 0.898 && - & - & \\
        && \textbf{\proposed{} (Ours)} & \textbf{34.90} & \textbf{0.900} && \textbf{37.46} & \textbf{0.924} & \\
        && \textbf{\proposed{}$\:\:$ $+$ $\rrpp$ (Ours)} & \textbf{35.97} & \textbf{0.925} && \textbf{38.09} & \textbf{0.937} & \\
        && \textbf{\proposed{}$^\benchmark$ $+$ $\rrpp$ (Ours)} & \textbf{36.91} & \textbf{0.931} && - & - & \\
        \bottomrule
    \end{tabularx}
    \vspace{-2mm}
    \caption{
        \textbf{Quantitative comparison of various denoising methods on the SIDD and DND benchmarks.}
        We note that several supervised methods leverage SIDD noisy-clean pairs for training and perform much better than our AP-BSN, while we use noisy sRGB images only for training.
        By default, we report official evaluation results from SIDD and DND benchmark websites.
        $\diamond$ and \textbf{R} indicate that the performances are evaluated by ourselves, or reported from R2R~\cite{R2R}, respectively.
        We also mark methods with $\ast$ which adopt self-ensemble strategy~\cite{EDSR}.
        $\benchmark$ denotes that the model is trained on SIDD benchmark images in a fully self-supervised fashion.
    }
    \label{tab:main}
    \vspace{-4mm}
\end{table*}

%% file: main/figs/fig_rrpp_exp.tex
\begin{figure}[t]
    \renewcommand{\wp}{0.49\linewidth}
    \renewcommand{\vs}{-2mm}
    \captionsetup[subfloat]{font=small}
    \centering
    \subfloat[Ablations on $p$]{\includegraphics[width=\wp]{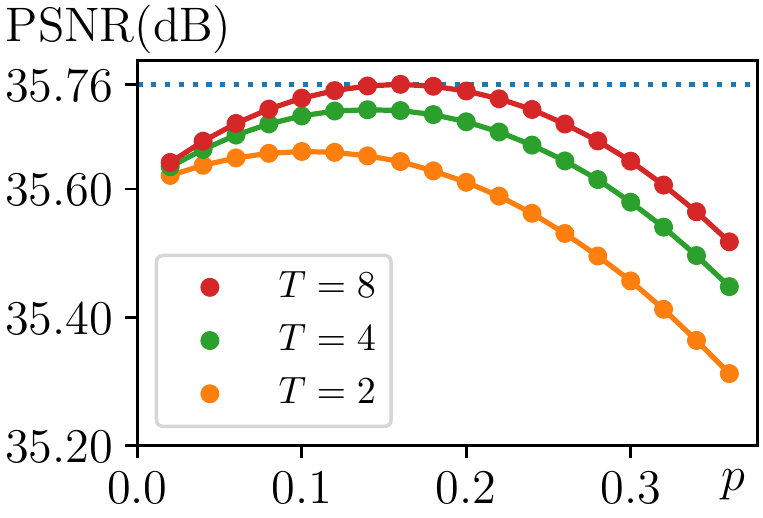}\label{fig:rrpp_exp_p}}
    \hfill
    \subfloat[Ablations on $T$]{\includegraphics[width=\wp]{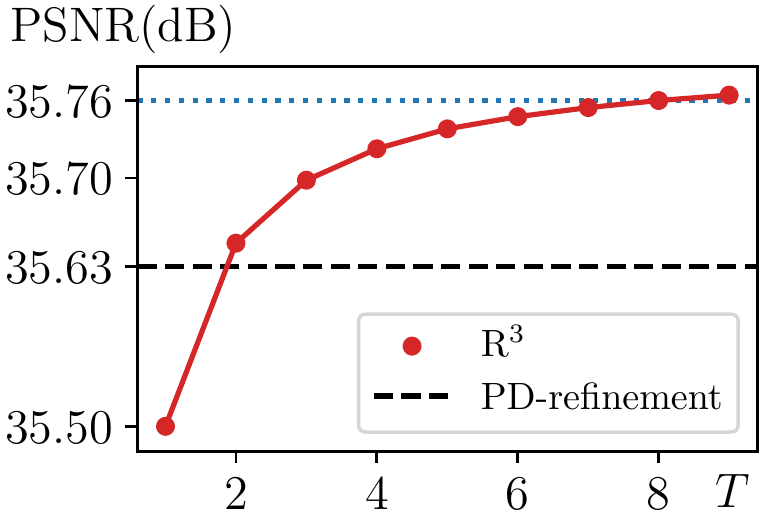}\label{fig:rrpp_exp_T}}
    \\
    \vspace{\vs}
    \caption{
    \textbf{Ablation study of AP-BSN + $\rrpp{}$ on the SIDD validation dataset.}
    We note that AP-BSN without $\rrpp{}$ achieves 34.86dB on the same dataset.
    %
    (a) We investigate the effect of different $p$ for $T = 2, 4, 8$.
    (b) We fix $p = 0.16$ to see the effect of $T$ in our $\rrpp{}$.
    }
    \label{fig:rrpp_exp}
    \vspace{-6mm}
\end{figure}

%% file: main/figs/fig_comparison.tex
\begin{figure*}[t]
    \renewcommand{\wp}{0.198 \linewidth}
    \captionsetup{position=top}
    \captionsetup[subfloat]{font=footnotesize, justification=centering}
    \centering
    \addtocounter{subfigure}{0}
    \subfloat{\includegraphics[width=\wp]{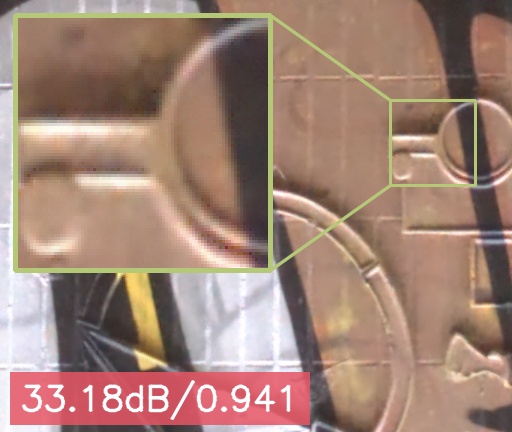}}
    \hfill
    \subfloat{\includegraphics[width=\wp]{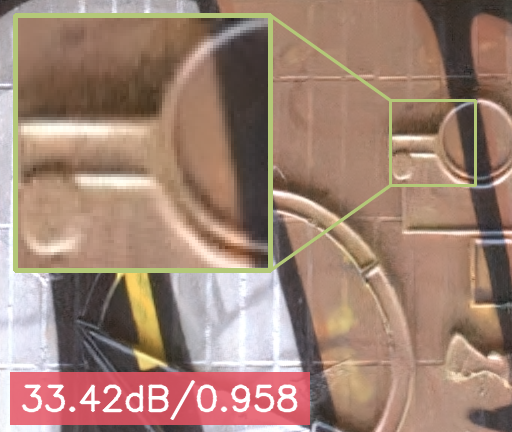}}
    \hfill
    \subfloat{\includegraphics[width=\wp]{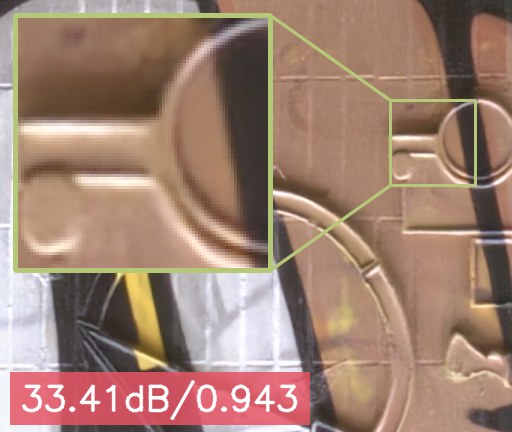}}
    \hfill
    %
    \subfloat[NAC~\cite{NAC}]{\includegraphics[width=\wp]{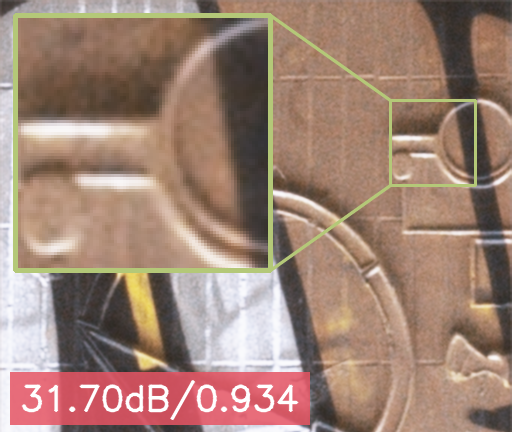}}
    \hfill
    \subfloat{\includegraphics[width=\wp]{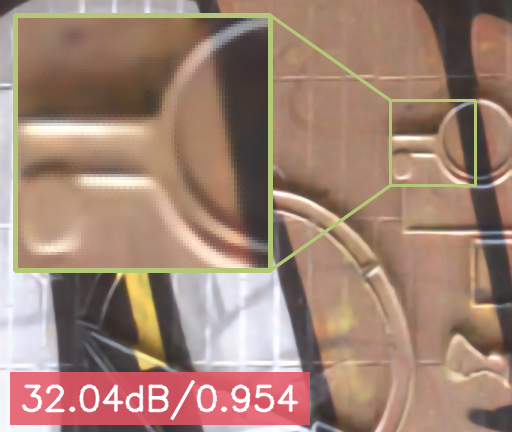}}
    \\
    %
    \addtocounter{subfigure}{-5}
    \captionsetup{position=bottom}
    \subfloat[DnCNN~\cite{DnCNN} \\ Supervised - Real SIDD ]{\includegraphics[width=\wp]{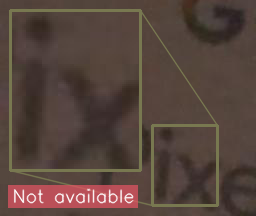}}
    \hfill
    \subfloat[Zhou~\etal~\cite{whenAWGN} \\ Supervised - Synthetic noise ]{\includegraphics[width=\wp]{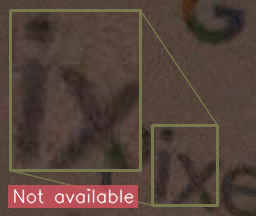}}
    \hfill
    \subfloat[C2N~\cite{C2N} $+$ DIDN~\cite{DIDN} \\ Unpaired ]{\includegraphics[width=\wp]{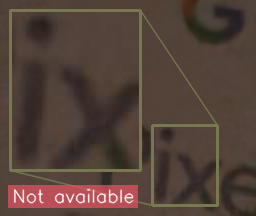}}
    \hfill
    %
    %
    \subfloat[R2R~\cite{R2R} \\ Self-supervised \addtocounter{subfigure}{1}]{\includegraphics[width=\wp]{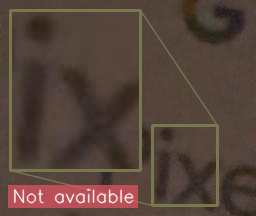}}
    %
    \hfill
    \subfloat[\textbf{\proposed{} + $\rrpp_{\time 8}$ (Ours)} \\ Self-supervised \addtocounter{subfigure}{-1}]{\includegraphics[width=\wp]{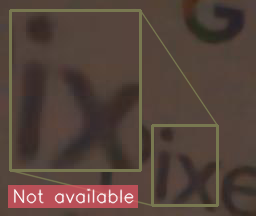}}
    %
    \\
    \vspace{-1mm}
    \caption{
    \textbf{Qualitative comparison between different denoising methods on DND~\cite{DND} and SIDD~\cite{SIDD} benchmarks.}
    %
    (a) DnCNN is trained on the paired SIDD-Medium dataset.
    %
    (b) Zhou~\etal train their method on synthetic AWGN and impulse noise.
    The learned denoising model is then combined with PD to handle real-world noise.
    (c) C2N generates a realistic noisy image from the clean input, where the following denoising model, \ie, DIDN, is trained on the generated pairs.
    %
    (d--e) Recent self-supervised approaches are trained on noisy images only.
    %
    (f) Our method is directly learnable on the practical sRGB images.
    We note that the DND benchmark (\textbf{Upper}) provides per-sample PSNR/SSIM, while SIDD benchmark (\textbf{Lower}) does not, \ie, Not available.
    }
    \label{fig:comparison}
    \vspace{-4mm}
\end{figure*}

%% file: main/6.tex
\section{Conclusion}
\label{sec:conclusion}
In this paper, we first identify several trade-offs regarding different PD stride factors in perspective of BSN.
Rather than directly integrate PD and BSN, we propose asymmetric PD between training and inference to satisfy pixel-wise independent assumption while preserving image details.
To this end, we propose AP-BSN, a fully self-supervised approaches for real-world denoising.
We also propose random-replacing refinement $\rrpp{}$, which removes visual artifacts of AP-BSN without any additional parameters.
The proposed AP-BSN + $\rrpp{}$ does not require any prior knowledge on real-world noise and outperforms recent self-supervised/unsupervised denoising methods.

%% file: supple/implementation.tex

\section{Optimization}
To train our \proposed{}, we randomly crop $120 \times 120$ noisy patches from the SIDD and DND datasets, respectively.
We note that 24,542, 24,784, and 24,320 patches are used in one epoch for SIDD-Medium, DND, and SIDD benchmark datasets, respectively.
Each sample is augmented with random $90^\circ$ rotation and horizontal/vertical flips.
Our mini-batch contains the 8 augmented samples.
The proposed \proposed{} is optimized for 20 epochs, where the learning rate is decayed by a factor of 10 for every 8 epochs.

\section{Network architecture}
Our BSN architecture is based on Wu~\etal~\cite{DBSN}, while several changes are made for simplification.
Instead of the MDC modules with multiple branches of the dilated convolutions, we use a sequence of dilated convolution modules (DC) that have a single branch only.
\figref{fig:network} visualizes a detailed architecture of the BSN used to construct our \proposed{} framework.

Therefore, our network has 3.7M parameters, which are fewer than 6.6M parameters from the original BSN proposed by Wu~\etal~\cite{DBSN}.
%
%
We also note that recent unsupervised/unpaired methods adopt larger denoising networks than the proposed AP-BSN.
Specifically, \eg DIDN~\cite{DIDN} and C2N~\cite{C2N},  MWCNN~\cite{MWCNN} has $\sim$16.2M in Wu~\etal~\cite{DBSN}).
Our AP-BSN w/o $\rrpp$ shows comparable results with much smaller denoising networks even our AP-BSN only uses noisy images.

%% file: supple/ablation.tex
\section{Effects of aliasing artifacts}
To examine the effect of aliasing artifact during the training and inference, we train our \proposed{} using clean SIDD images \emph{only}.
Specifically, BSN is trained to reconstruct the same image from given a clean input while not seeing the center pixel in the receptive field.
We suppose that the clean images contain zero-intensity noise, which follows the two basic assumptions of BSN: noise signals are spatially uncorrelated and zero-mean.
Thus, PD-BSN should learn an identity mapping if sub-images from PD do not contain any noise.
However, as shown in Figs.~\ref{fig:c2c_22} and \ref{fig:c2c_55}, $\pd{5}$-BSN removes high-frequency information from the given input clean image in \figref{fig:c2c_clean} and does not operate an identity function even on the clean image while $\pd{2}$-BSN does not.
From this observation, we can assume that $\pd{5}$-BSN learns to remove some information during the training that does not exist in $\pd{2}$ sub-images.
When we apply the proposed $\ap{5}{2}$ strategy, BSN does not remove high-frequency components and preserves the image structure well, as shown in \figref{fig:c2c_52}.
Therefore, we conclude that the aliasing artifacts prevent $\pd{5}$-BSN from being a feasible denoising model since removing the artifacts during inference can significantly degrade the performance of PD-BSN.

\input{supple/figs/fig_c2c}

%% file: supple/figs/fig_c2c.tex
\begin{figure}[t!]
    \renewcommand{\wp}{0.495\linewidth}
    \renewcommand{\vs}{-2mm}
    \captionsetup[subfloat]{font=small}
    \centering
    \subfloat[Input clean image $\img{C}$]
    {\includegraphics[width=\wp]{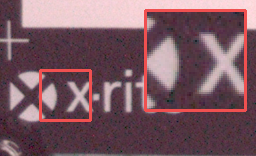}\label{fig:c2c_clean}}
    \hfill
    \subfloat[$\pd{2}$-BSN]
    {\includegraphics[width=\wp]{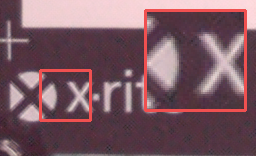}\label{fig:c2c_22}}
    \\
    \subfloat[$\pd{5}$-BSN]
    {\includegraphics[width=\wp]{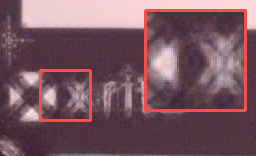}\label{fig:c2c_55}}
    \hfill
    \subfloat[\textbf{$\ap{5}{2}$-BSN (Ours)}]
    {\includegraphics[width=\wp]{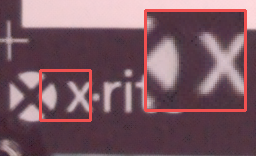}\label{fig:c2c_52}}
    \vspace{-2mm}
    \caption{
        \textbf{Effects of aliasing artifacts in BSN.}
        To validate that the advantage of AP-BSN comes from the existence of aliasing artifacts, we conduct a clean-to-clean experiment.
        We sample a clean image from the SIDD validation dataset for visualization.
    }
    \label{fig:c2c}
    \vspace{-4mm}
\end{figure}

%% file: supple/nind.tex
\input{supple/figs/fig_nind}

\section{AP-BSN on the NIND dataset}
In Fig.~\textcolor{red}{2a} of our main manuscript, we have demonstrated that noise signals in the NIND~\cite{NIND} dataset show gradually decreasing correlations between them as their relative distance $d$ increases.
Such observation implies that the proposed \proposed{} may perform better with $a = 6$ or larger, as the spatial correlations between noise can be further reduced.
Therefore, we analyze the trade-offs of $\ap{a}{b}$ on the NIND dataset similar to Section \textcolor{red}{5.2} in our main manuscript.
To investigate the trade-off under diverse scenes, we conduct a per-sample analysis rather than calculating the performance on the entire dataset.
\figref{fig:nind_noisy} shows several noisy images in the NIND dataset.
In \figref{fig:nind_denoised}, we also visualize the denoising results of our $\ap{5}{2}$-BSN~$+$~$\rrpp{}$ trained on the NIND dataset.
Since the noise property of the NIND dataset differs from SIDD, $\ap{6}{2}$ may perform slightly better on some specific samples as shown in \figref{fig:nind_tradeoff}.
However, we note that the performance gaps are marginal, and $\ap{5}{2}$ generalizes well on various real-world datasets on average.
%

%% file: supple/figs/fig_nind.tex
\begin{figure*}[t]
    \renewcommand{\wp}{0.27\linewidth}
    \newcommand{\hp}{0.23\linewidth}
    \newcommand{\chp}{0.23\linewidth}
    \newcommand{\mhp}{1mm}
    \newcommand{\rhp}{8mm}
    \newcommand{\hhp}{3mm}
    \captionsetup[subfloat]{font=small, justification=centering}
    \centering
    \hfill 
    \subfloat{\includegraphics[height=\hp]{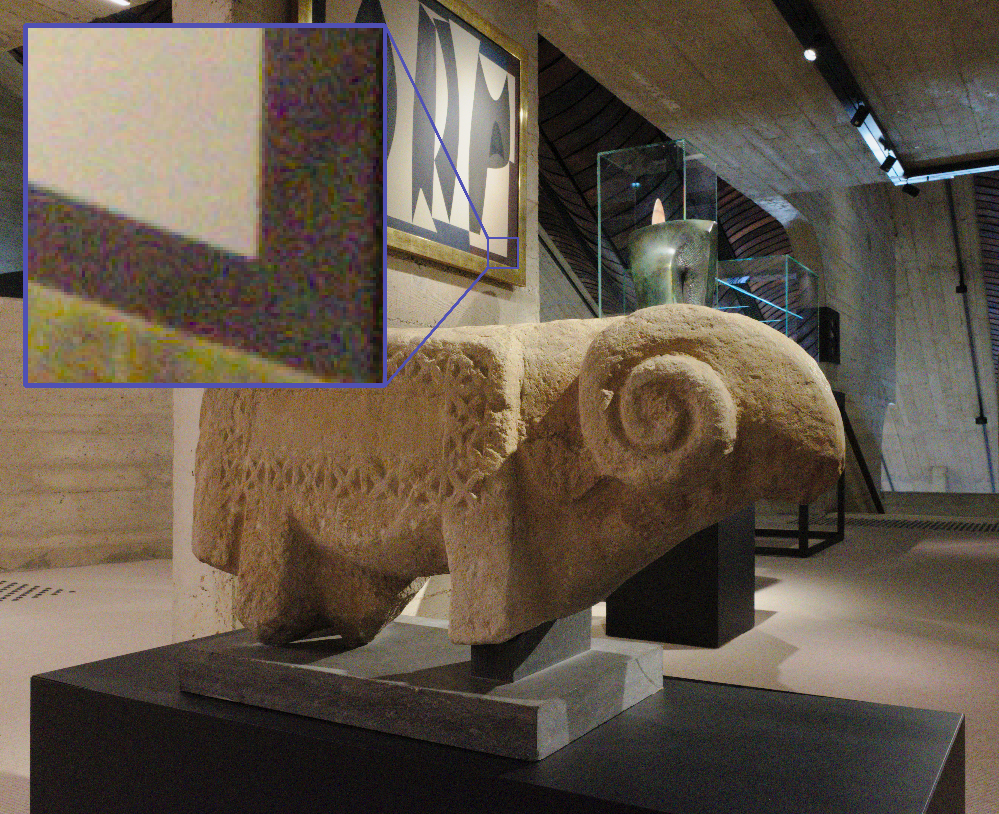}}
    \hspace{\mhp}
    \subfloat{\includegraphics[height=\chp]{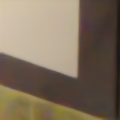}}
    \hspace{\hhp}
    \subfloat{\includegraphics[width=\wp]{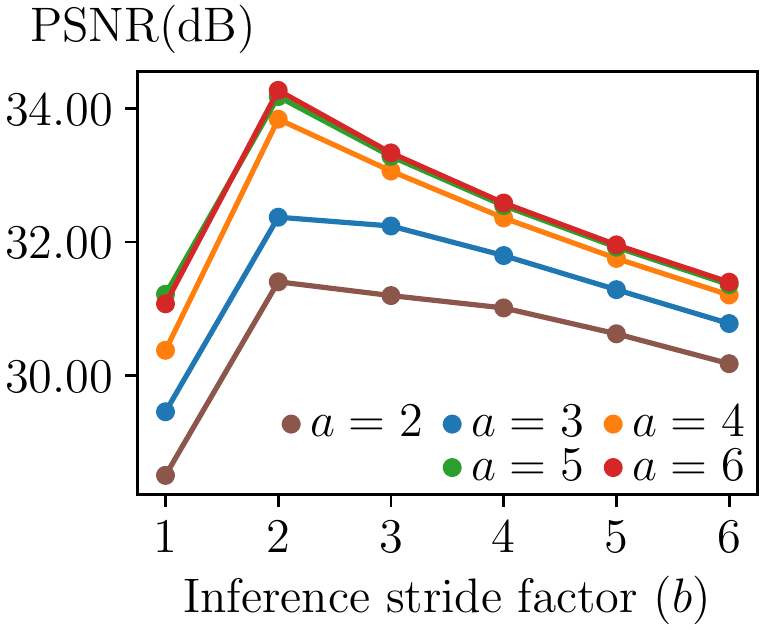}}
    \hspace{\rhp} \null
    \\
    \vspace{1mm}
    \hfill 
    \subfloat{\includegraphics[height=\hp]{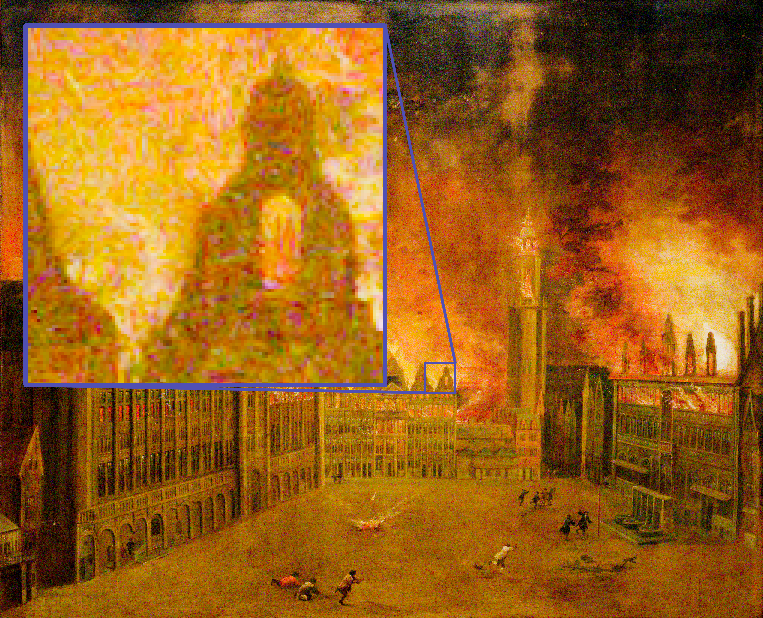}}
    \hspace{\mhp}    
    \subfloat{\includegraphics[height=\chp]{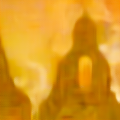}}
    \hspace{\hhp}
    \subfloat{\includegraphics[width=\wp]{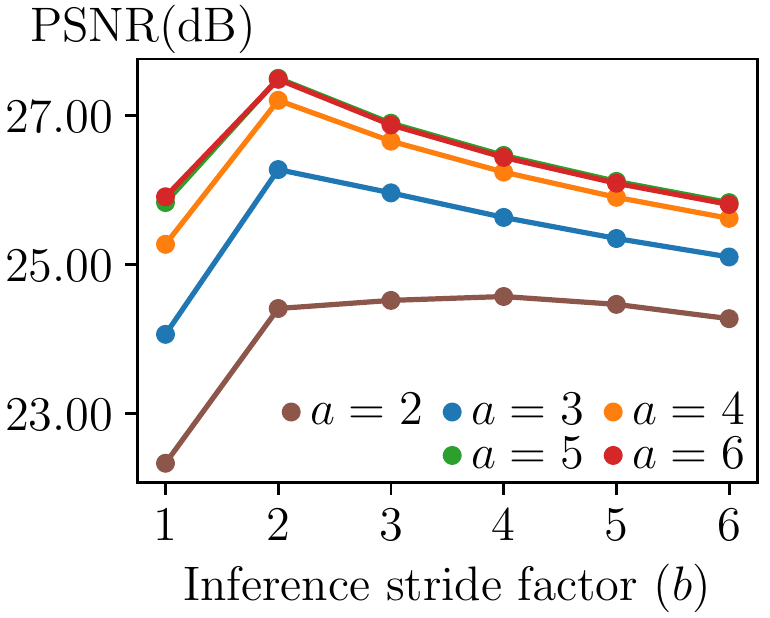}}
    \hspace{\rhp} \null
    \\
    \vspace{1mm}
    \hfill 
    \subfloat{\includegraphics[height=\hp]{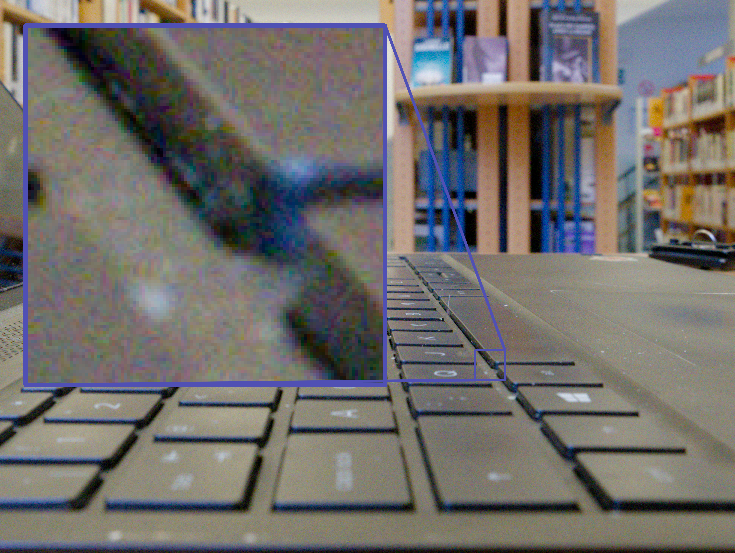}}
    \hspace{\mhp}
    \subfloat{\includegraphics[height=\chp]{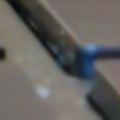}}
    \hspace{\hhp}
    \subfloat{\includegraphics[width=\wp]{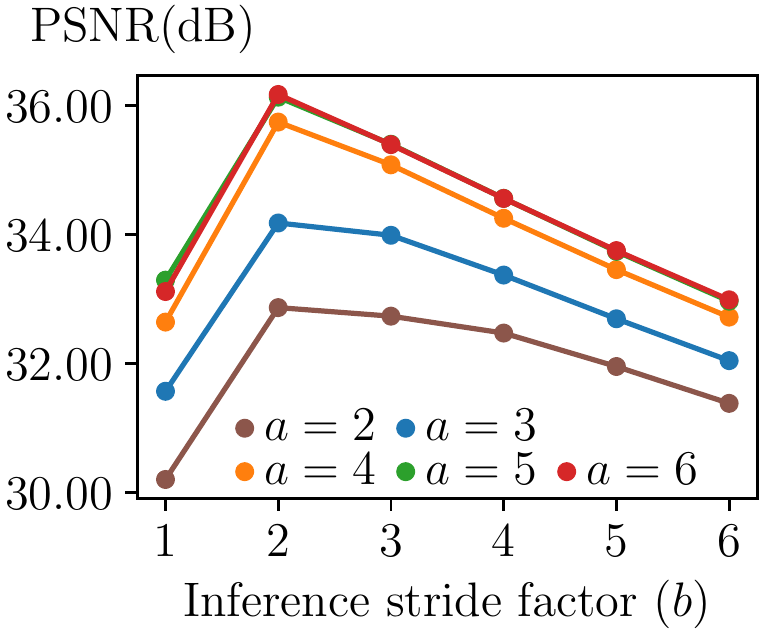}}
    \hspace{\rhp} \null
    \\
    \vspace{1mm}
    \hfill 
    \subfloat{\includegraphics[height=\hp]{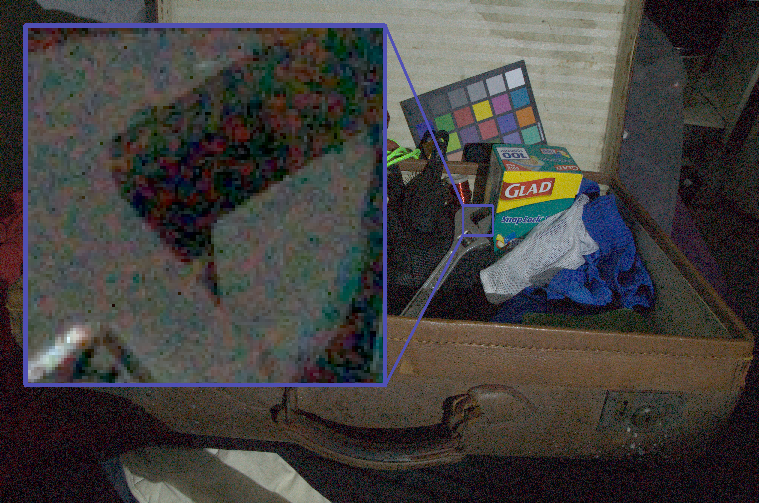}}
    \hspace{\mhp} 
    \subfloat{\includegraphics[height=\chp]{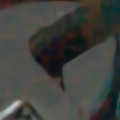}}
    \hspace{\hhp}
    \subfloat{\includegraphics[width=\wp]{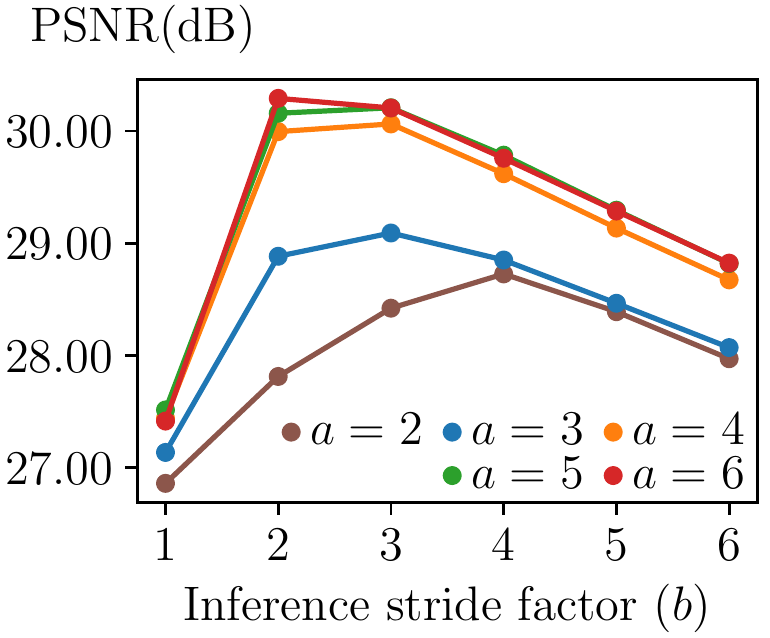}}
    \hspace{\rhp} \null
    \\
    \addtocounter{subfigure}{-12}
    \vspace{1mm}
    \hfill
    \subfloat[Noisy images from the NIND~\cite{NIND} dataset \label{fig:nind_noisy}] {\includegraphics[height=\hp]{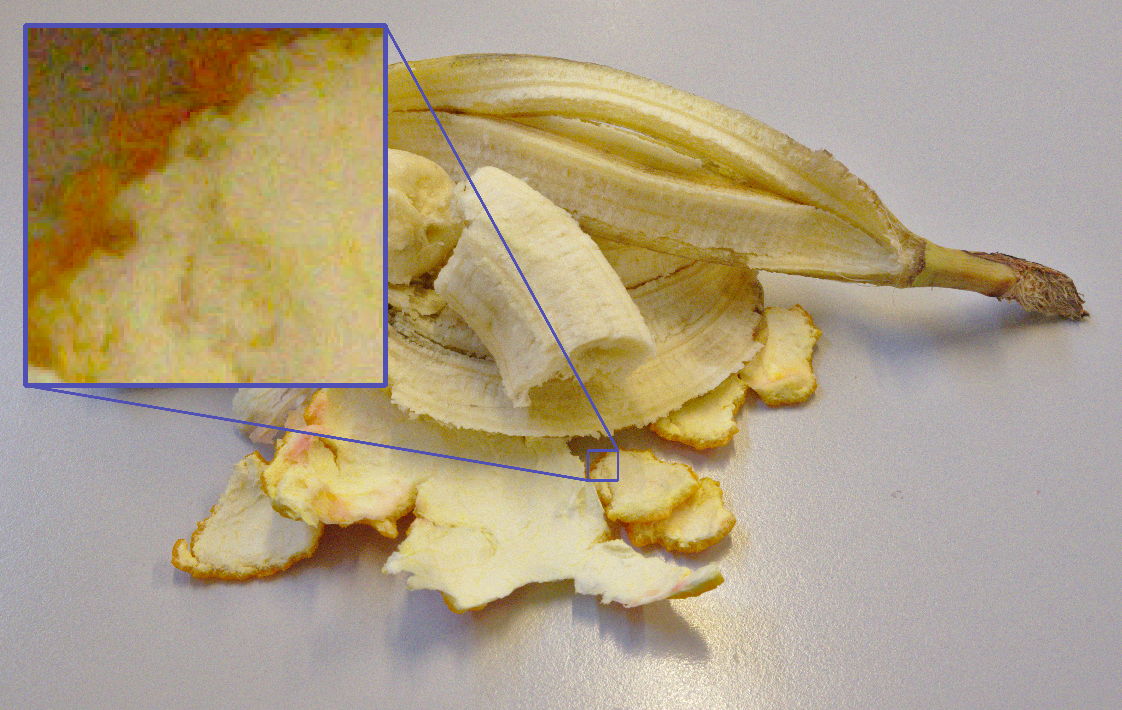}}
    \hspace{\mhp}         
    \subfloat[\textbf{\proposed{}~$+$~$\rrpp{}$ (Ours)} \label{fig:nind_denoised}]{\includegraphics[height=\chp]{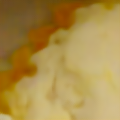}}
    \hspace{\hhp}
    \subfloat[Effects of $\ap{a}{b}$ \label{fig:nind_tradeoff}]{\includegraphics[width=\wp]{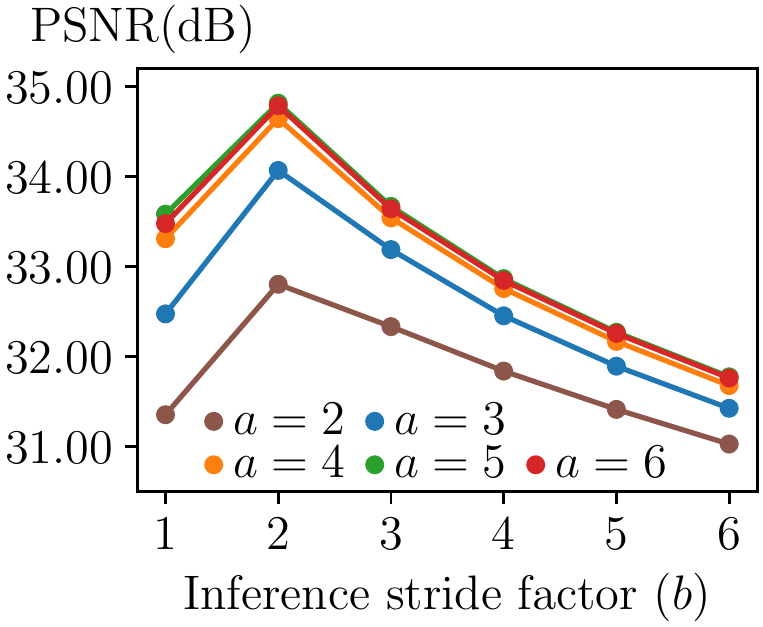}}
    \hspace{\rhp} \null
    \\
    \vspace{-2mm}
    \caption{
        \textbf{Per-sample analysis of $\ap{a}{b}$ on the NIND dataset.}
        (a) Noisy images sampled from the NIND dataset.
        From top: `NIND\_MuseeL-ram\_ISO6400.jpg,' `NIND\_MVB-Bombardement\_ISOH1.jpg,' `NIND\_LaptopInLibrary\_ISO2500.png,' `NIND\_Iain02\_ISO3200.png,' `NIND\_partiallyeatenbanana\_ISO2500.png.'
        (b) Results of our \proposed{} + $\rrpp{}$ on the NIND dataset.
        We show local patches for better visualization.
        (c) Per-image trade-off analysis.
        The proposed AP-BSN performs consistently well when $b = 2$, while the best performance can be achieved when the training stride factor $a$ is set to 5 or 6.
        Please see Fig.~\textcolor{red}{7} in our main manuscript for more details.
    }
    \label{fig:nind}
    \vspace{-4mm}
\end{figure*}

%% file: supple/more.tex
\section{Qualitative results}

\subsection{Additional qualitative results}
Since several existing methods do not provide qualitative results on specific datasets, we could not perform extensive qualitative comparisons in our main manuscript.
For example, Figs.~\textcolor{red}{10d} (upper figure in the 3rd column) and \textcolor{red}{10e} (lower figure in the 3rd column) in our main manuscript represent results of NAC~\cite{NAC} on the DND benchmark and R2R~\cite{R2R} on the SIDD benchmark, respectively, because R2R does not provide results on the DND dataset.
\figref{fig:more} shows additional qualitative comparison between different denoising methods on the DND~\cite{DND} benchmark and SIDD~\cite{SIDD} validation dataset.

\subsection{Results on real-world inputs}
Our \proposed{} is designed to handle real-world sRGB images, where appropriate training examples, \ie, noisy-real pairs for supervised, a set of clean images for unpaired learning, may not exist.
One of the major advantages of the proposed fully self-supervised framework is that we can apply our model on a \emph{single} noisy test image directly without any pre-trained knowledge.
To this end, we capture real-world noisy images under a high ISO condition using the recent Samsung Galaxy smartphone.
Modern smartphone cameras usually incorporate software-based denoising algorithms to remove unpleasing noise from the captured scene.
Therefore, we first acquire RAW data and leverage the simulated camera pipeline without explicit denoising stage~\cite{SIDD} to get the corresponding sRGB images.

\figref{fig:ryan} visualizes denoising results of our method on the real-world sRGB images.
Compared to the hardware-specific in-camera denoising algorithm in \figref{fig:ryan_incamera}, our approach reconstructs much sharper edges while suppressing unwanted noise signals effectively, as shown in \figref{fig:ryan_ours}.
The proposed method also outperforms DnCNN~\cite{DnCNN} trained on SIDD~\cite{SIDD} noisy-clean pairs, while our formulation utilizes a \emph{single} noisy image only for training.

\subsection{Qualitative improvement by $\rrpp$}
Our $\rrpp$ post-processing strategy significantly improves the performance of the proposed denoising method.
\figref{fig:rrpp_qualitative} provides qualitative comparisons between AP-BSN \emph{without} $\rrpp$ and AP-BSN~$+$~$\rrpp$.
Without $\rrpp{}$, our AP-BSN tends to generate unpleasing blocky artifacts as shown in \figref{fig:rq4ap}.
By using the proposed $\rrpp{}$, our AP-BSN can reconstruct smooth and natural image structures without requiring any additional parameters and training.


\input{supple/figs/fig_more}
\clearpage

\input{supple/figs/fig_ryan}
\clearpage

\input{supple/figs/fig_rrpp_qualitative}
\clearpage

%% file: supple/figs/fig_more.tex
\begin{figure*}[t]
    \renewcommand{\wp}{0.195 \linewidth}
    \captionsetup[subfloat]{font=footnotesize, justification=centering}
    \centering
    \subfloat{\includegraphics[width=\wp]{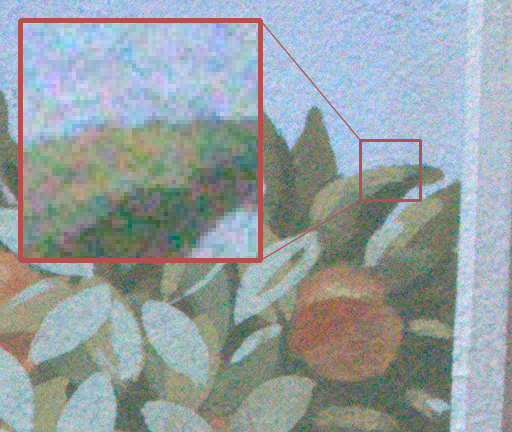}}
    \hfill
    \subfloat{\includegraphics[width=\wp]{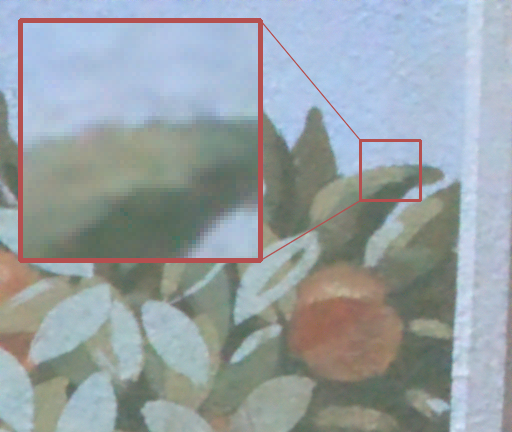}}
    \hfill
    \subfloat{\includegraphics[width=\wp]{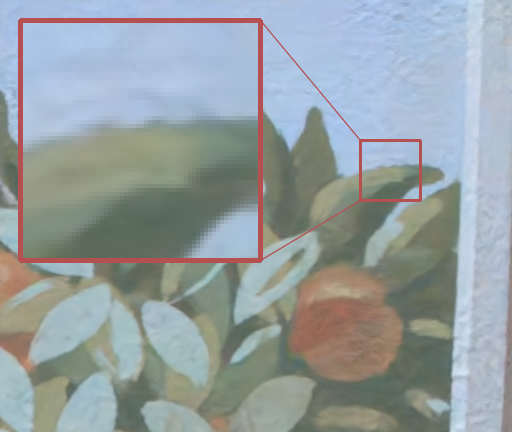}}
    \hfill
    \subfloat{\includegraphics[width=\wp]{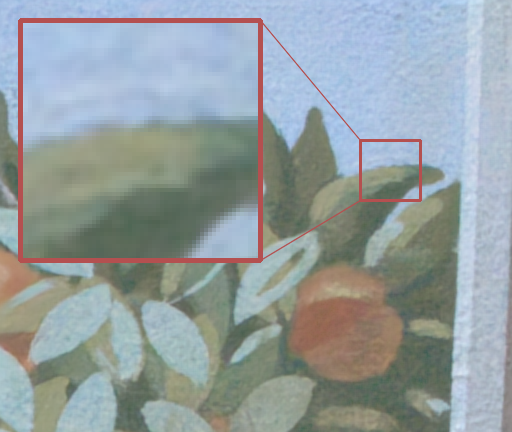}}
    \hfill
    \subfloat{\includegraphics[width=\wp]{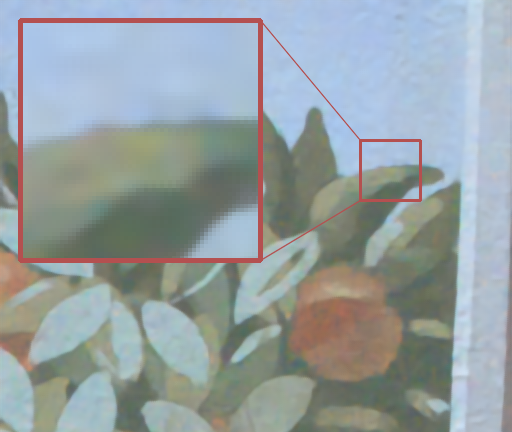}}
    \\
    \captionsetup[subfigure]{labelformat=empty}
    \addtocounter{subfigure}{-5}
    \vspace{0.5mm}
    \subfloat{\includegraphics[width=\wp]{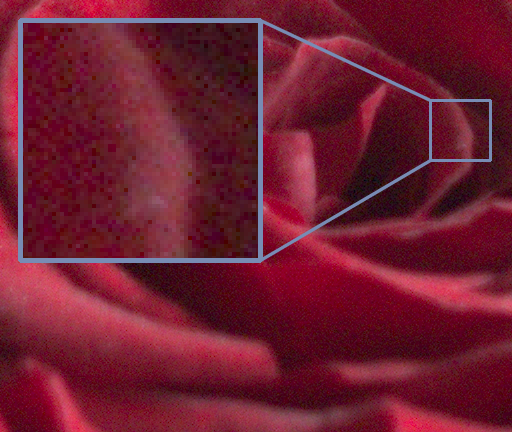}}
    \hfill
    \subfloat[\normalsize{DND~\cite{DND} dataset $\uparrow$}]{\includegraphics[width=\wp]{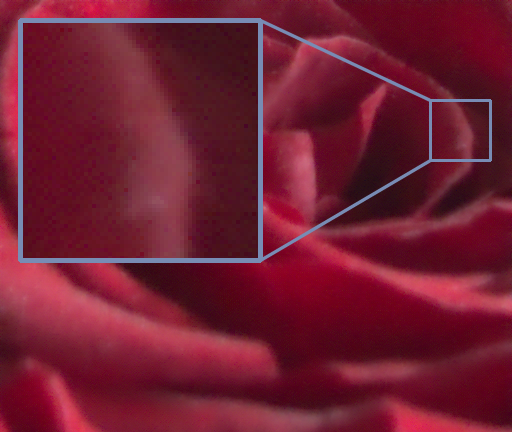}}
    \hfill
    \subfloat{\includegraphics[width=\wp]{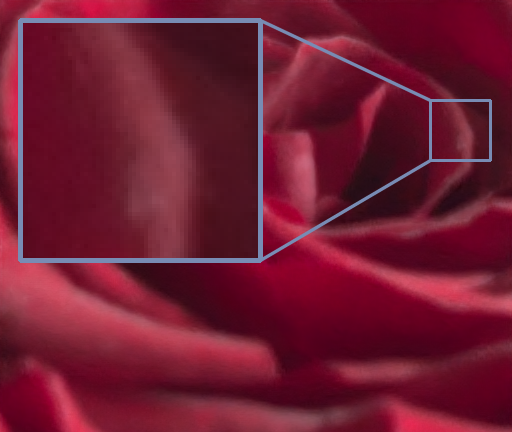}}
    \hfill
    \subfloat[\normalsize{$\downarrow$ SIDD~\cite{SIDD} dataset}]{\includegraphics[width=\wp]{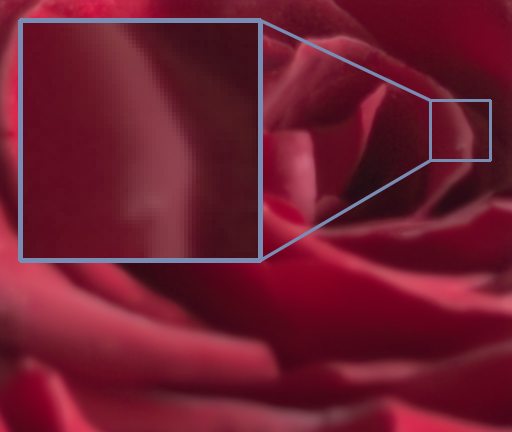}}
    \hfill
    \subfloat{\includegraphics[width=\wp]{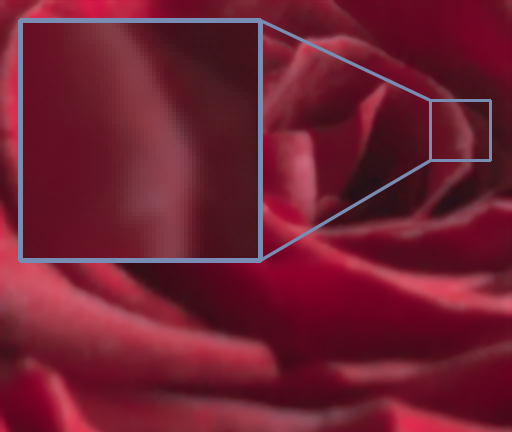}}
    \\
    \addtocounter{subfigure}{-5}
    \subfloat{\includegraphics[width=\wp]{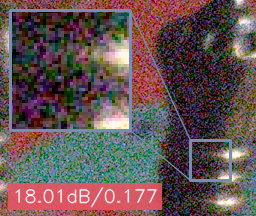}}
    \hfill
    \subfloat{\includegraphics[width=\wp]{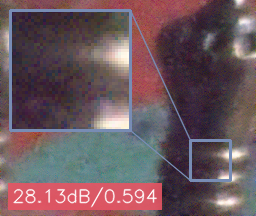}}
    \hfill
    \subfloat{\includegraphics[width=\wp]{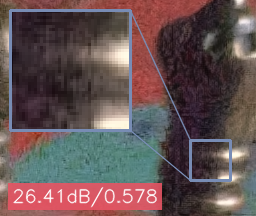}}
    \hfill
    \subfloat{\includegraphics[width=\wp]{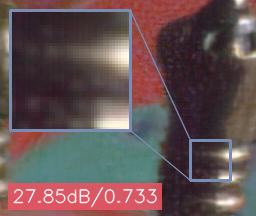}}
    \hfill
    \subfloat{\includegraphics[width=\wp]{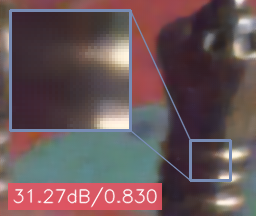}}
    \\
    \addtocounter{subfigure}{-5}
    \vspace{0.5mm}
    \subfloat{\includegraphics[width=\wp]{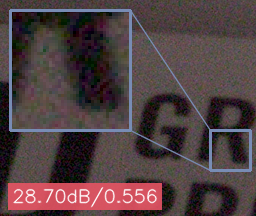}}
    \hfill
    \subfloat{\includegraphics[width=\wp]{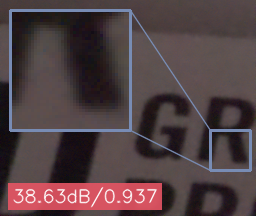}}
    \hfill
    \subfloat{\includegraphics[width=\wp]{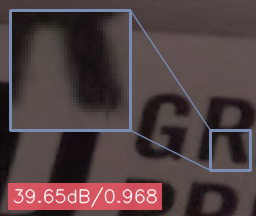}}
    \hfill
    \subfloat{\includegraphics[width=\wp]{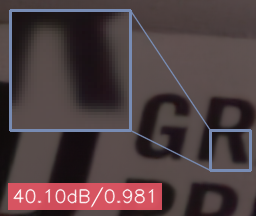}}
    \hfill
    \subfloat{\includegraphics[width=\wp]{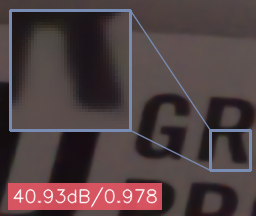}}
    \\
    \addtocounter{subfigure}{-5}
    \vspace{0.5mm}
    \subfloat{\includegraphics[width=\wp]{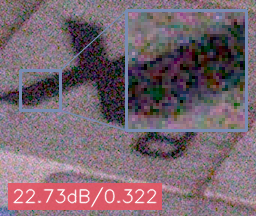}}
    \hfill
    \subfloat{\includegraphics[width=\wp]{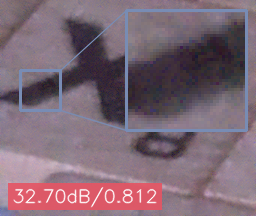}}
    \hfill
    \subfloat{\includegraphics[width=\wp]{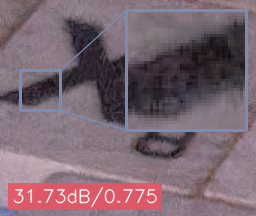}}
    \hfill
    \subfloat{\includegraphics[width=\wp]{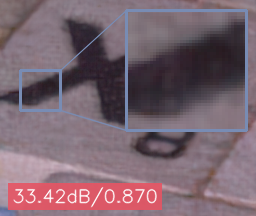}}
    \hfill
    \subfloat{\includegraphics[width=\wp]{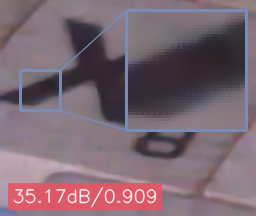}}
    \\
    \captionsetup[subfigure]{labelformat=parens}
    \addtocounter{subfigure}{-5}
    \vspace{0.5mm}
    \subfloat[Noisy]{\includegraphics[width=\wp]{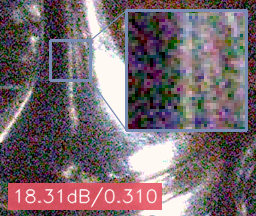}}
    \hfill
    \subfloat[DnCNN~\cite{DnCNN} \\ Supervised - Real SIDD]{\includegraphics[width=\wp]{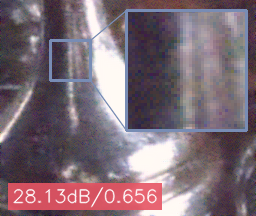}}
    \hfill
    \subfloat[Zhou~\etal~\cite{whenAWGN} \\ Supervised - Synthetic noise]{\includegraphics[width=\wp]{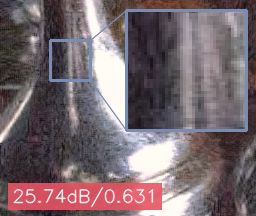}}
    \hfill
    \subfloat[C2N~\cite{C2N} $+$ DIDN~\cite{DIDN} \\ Unpaired]{\includegraphics[width=\wp]{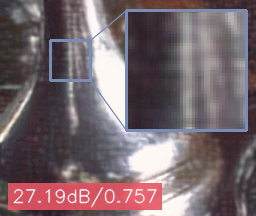}}
    \hfill
    \subfloat[\textbf{\proposed{} + $\rrpp_{\time 8}$ (Ours)} \\ Self-supervised]{\includegraphics[width=\wp]{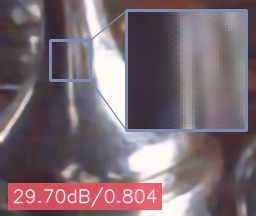}}
    \\
    \vspace{-2mm}
    \caption{
    \textbf{Additional qualitative comparison between different methods on DND~\cite{DND} benchmark and SIDD~\cite{SIDD} validation datasets.}
    The upper two rows are examples from the DND benchmark dataset, and the lower four rows are from the SIDD validation dataset.
    (a) Input noisy images.
    (b) Same as Fig.~\textcolor{red}{10a} in our main manuscript, DnCNN is trained on the paired SIDD-Medium dataset.
    (c) Zhou~\etal train their method on synthetic AWGN and impulse noise.
    During the inference, $\pd{2}$ is used to break the spatial correlation of real-world noise.
    (d) C2N generates a realistic noisy image from the clean input, where the following denoising model, \ie, DIDN, is trained on the generated pairs.
    (e) Our method is directly applicable to practical sRGB noisy images in a self-supervised manner, which does not require any additional data.
    For quantitative comparison, we mark per-sample PSNR/SSIM \wrt the ground-truth image at the bottom left of each patch.
    We also note that ground-truth images are not available for the DND dataset.
    }
    \label{fig:more}
    \vspace{-4mm}
\end{figure*}

%% file: supple/figs/fig_ryan.tex
\begin{figure*}[t!]
    \renewcommand{\wp}{0.1715 \linewidth}
    \newcommand{\pad}{0.001 \linewidth}
    \renewcommand{\vs}{-2mm}
    \newcommand{\vvs}{4mm}
    \captionsetup[subfloat]{font=scriptsize}
    \centering
    \begin{minipage}{0.46 \linewidth}
        \subfloat{\includegraphics[width=\linewidth]{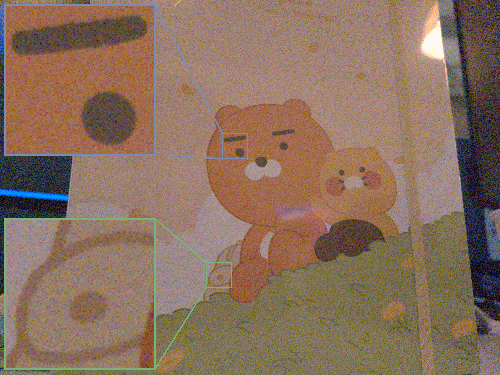}\label{fig:ryan_noisy}} \\
        \subfloat{\includegraphics[height=\pad]{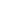}} \vspace{\vs} \\
        \subfloat{\includegraphics[width=\linewidth]{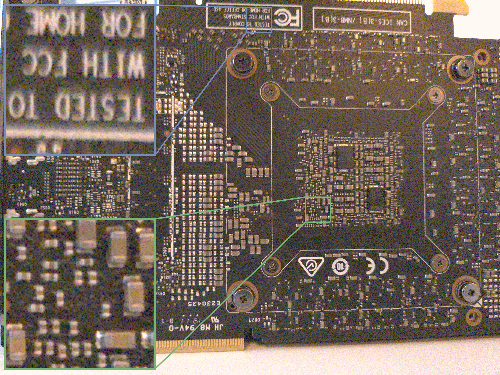}\label{fig:circuit_noisy}} \\
        \subfloat{\includegraphics[height=\pad]{_src/supple/ryan/pad.png}} \vspace{\vs} \\
        \addtocounter{subfigure}{-4}
        \subfloat[Real-world sRGB images under the high ISO condition ]{\includegraphics[width=\linewidth]{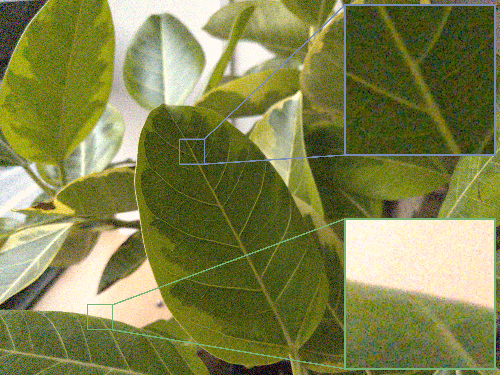}\label{fig:plant_noisy}}
    \end{minipage}
    \hfill
    \begin{minipage}{\wp}
        \subfloat{\includegraphics[width=\linewidth]{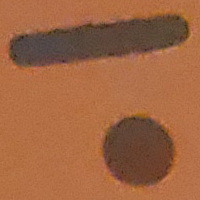}} \\
        \subfloat{\includegraphics[width=\linewidth]{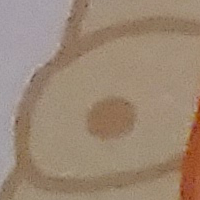}} \\
        \subfloat{\includegraphics[height=\pad]{_src/supple/ryan/pad.png}} \vspace{\vs} \\
        \subfloat{\includegraphics[width=\linewidth]{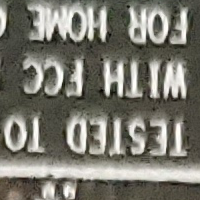}} \\
        \subfloat{\includegraphics[width=\linewidth]{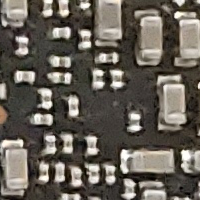}} \\
        \subfloat{\includegraphics[height=\pad]{_src/supple/ryan/pad.png}} \vspace{\vs} \\
        \subfloat{\includegraphics[width=\linewidth]{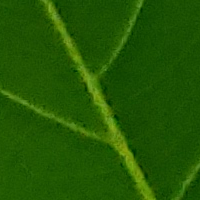}} \\
        \addtocounter{subfigure}{-7}
        \subfloat[In-camera processing ]{\includegraphics[width=\linewidth]{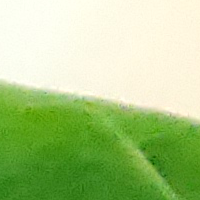}\label{fig:ryan_incamera}}
    \end{minipage}
    \hfill
    \begin{minipage}{\wp}
        \subfloat{\includegraphics[width=\linewidth]{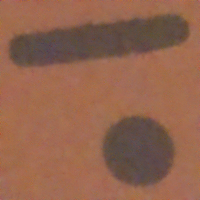}} \\
        \subfloat{\includegraphics[width=\linewidth]{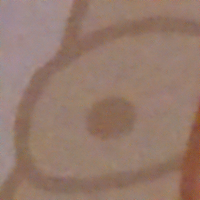}} \\
        \subfloat{\includegraphics[height=\pad]{_src/supple/ryan/pad.png}} \vspace{\vs} \\
        \subfloat{\includegraphics[width=\linewidth]{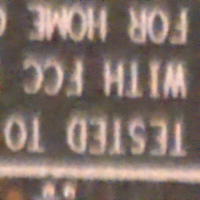}} \\
        \subfloat{\includegraphics[width=\linewidth]{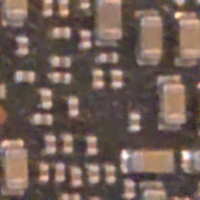}} \\
        \subfloat{\includegraphics[height=\pad]{_src/supple/ryan/pad.png}} \vspace{\vs} \\
        \subfloat{\includegraphics[width=\linewidth]{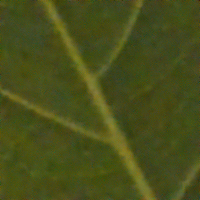}} \\
        \addtocounter{subfigure}{-7}
        \subfloat[DnCNN~\cite{DnCNN} on SIDD~\cite{SIDD} ]{\includegraphics[width=\linewidth]{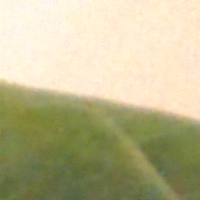}\label{fig:ryan_dncnn}}
    \end{minipage}
    \hfill
    \begin{minipage}{\wp}
        \vspace{0.2mm}
        \subfloat{\includegraphics[width=\linewidth]{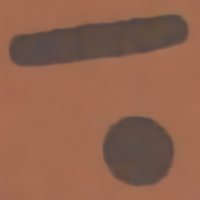}} \\
        \subfloat{\includegraphics[width=\linewidth]{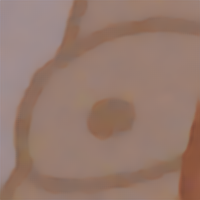}} \\
        \subfloat{\includegraphics[height=\pad]{_src/supple/ryan/pad.png}} \vspace{\vs} \\
        \subfloat{\includegraphics[width=\linewidth]{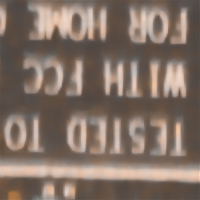}} \\
        \subfloat{\includegraphics[width=\linewidth]{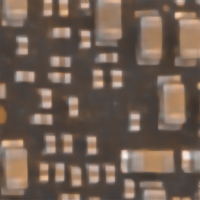}} \\
        \subfloat{\includegraphics[height=\pad]{_src/supple/ryan/pad.png}} \vspace{\vs} \\
        \subfloat{\includegraphics[width=\linewidth]{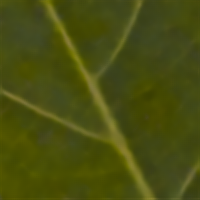}} \\
        \addtocounter{subfigure}{-7}
        \subfloat[\textbf{AP-BSN $+$ $\rrpp$ (Ours)} ]{\includegraphics[width=\linewidth]{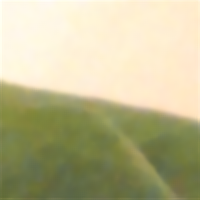}\label{fig:ryan_ours}}
    \end{minipage}
    \\
    \vspace{-2mm}
    \caption{
        \textbf{AP-BSN + $\text{R}^3$ on noisy images captured by ourselves.}
        (a) To avoid the in-camera denoising pipeline, we first capture RAW images with ISO 3200 using a recent Samsung Galaxy smartphone.
        We note that no other pre/post-processing is done on the exported \emph{.dng} files.
        Then, we render the sRGB images using the SIDD ISP pipeline~\cite{SIDD}, which does not include the denoising process.
        (b) The corresponding sRGB images processed by the smartphone.
        We note that the recent mobile devices have adopted software-based denoising algorithms, which suppress unwanted noise from the captured images.
        (c) Same as Fig.~\textcolor{red}{10a} in our main manuscript, DnCNN is trained on the real-world SIDD pairs.
        (d) Results of our AP-BSN + $\text{R}^3$ trained on a \emph{single} noisy input without any external data.
        We note that there exist color shifts between (a) and (b) since the simulated ISP pipeline does not know the color mappings of the actual ISP.
    }
    \label{fig:ryan}
    \vspace{-4mm}
\end{figure*}

%% file: supple/figs/fig_rrpp_qualitative.tex
\begin{figure*}[t!]
    \renewcommand{\wp}{0.245\linewidth}
    \renewcommand{\vs}{-2mm}
    \captionsetup[subfloat]{font=small}
    \centering
    \hfill
    \subfloat
    {\includegraphics[width=\wp]{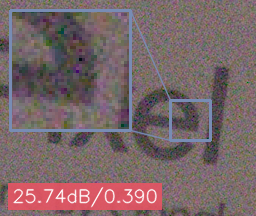}\label{fig:rq1n}}
    \hfill
    \subfloat
    {\includegraphics[width=\wp]{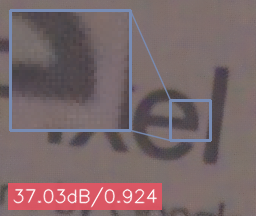}\label{fig:rq1ap}}
    \hfill
    \subfloat
    {\includegraphics[width=\wp]{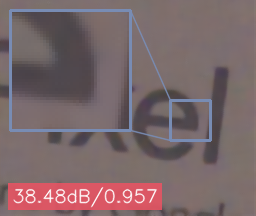}\label{fig:rq1apr}}
    \hfill
    \subfloat
    {\includegraphics[width=\wp]{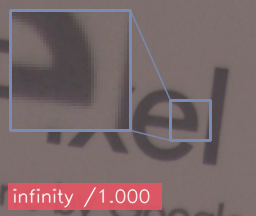}\label{fig:rq1c}}
    \hfill \null
    \\
    \hfill
    \subfloat
    {\includegraphics[width=\wp]{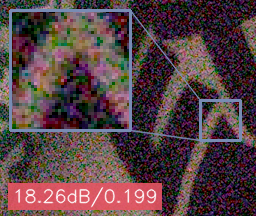}\label{fig:rq2n}}
    \hfill
    \subfloat
    {\includegraphics[width=\wp]{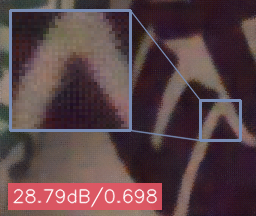}\label{fig:rq2ap}}
    \hfill
    \subfloat
    {\includegraphics[width=\wp]{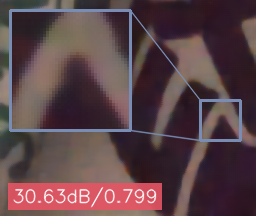}\label{fig:rq2apr}}
    \hfill
    \subfloat
    {\includegraphics[width=\wp]{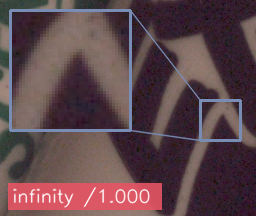}\label{fig:rq2c}}
    \hfill \null
    \\
    \hfill
    \subfloat
    {\includegraphics[width=\wp]{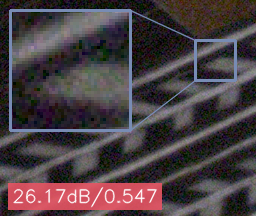}\label{fig:rq5n}}
    \hfill
    \subfloat
    {\includegraphics[width=\wp]{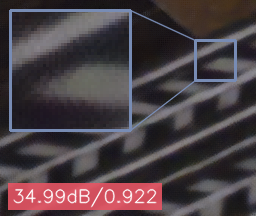}\label{fig:rq5ap}}
    \hfill
    \subfloat
    {\includegraphics[width=\wp]{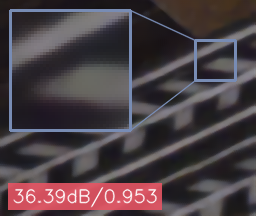}\label{fig:rq5apr}}
    \hfill
    \subfloat
    {\includegraphics[width=\wp]{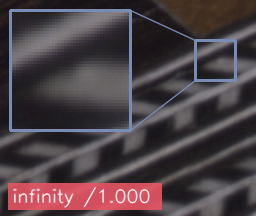}\label{fig:rq5c}}
    \hfill \null
    \\
    \hfill
    \subfloat
    {\includegraphics[width=\wp]{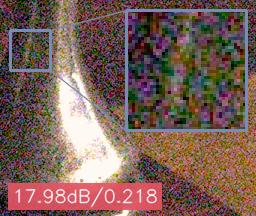}\label{fig:rq3n}}
    \hfill
    \subfloat
    {\includegraphics[width=\wp]{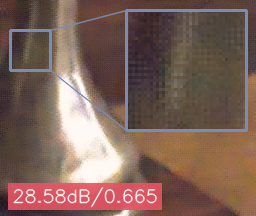}\label{fig:rq3ap}}
    \hfill
    \subfloat
    {\includegraphics[width=\wp]{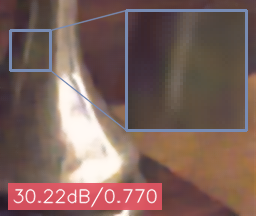}\label{fig:rq3apr}}
    \hfill
    \subfloat
    {\includegraphics[width=\wp]{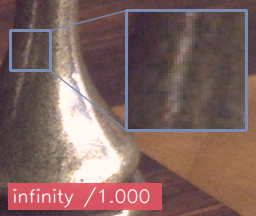}\label{fig:rq3c}}
    \hfill \null
    \\
    \addtocounter{subfigure}{-16}
    \hfill
    \subfloat[Noisy image $\img{N}$]
    {\includegraphics[width=\wp]{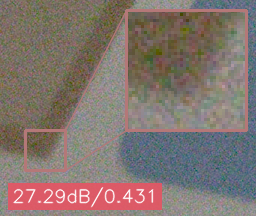}\label{fig:rq4n}}
    \hfill
    \subfloat[AP-BSN]
    {\includegraphics[width=\wp]{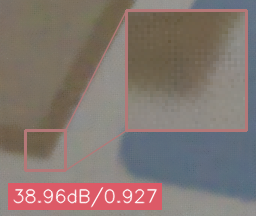}\label{fig:rq4ap}}
    \hfill
    \subfloat[\textbf{AP-BSN $+$ $\rrpp$ (Ours)}]
    {\includegraphics[width=\wp]{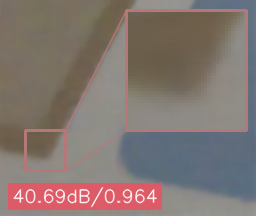}\label{fig:rq4apr}}
    \hfill
    \subfloat[Clean image $\img{C}$]
    {\includegraphics[width=\wp]{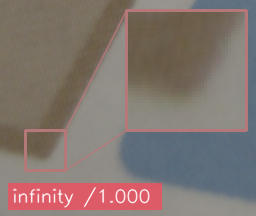}\label{fig:rq4c}}
    \hfill \null
    \\
    \vspace{-2mm}
    \caption{
        \textbf{Visual comparison between denoising results of AP-BSN \emph{without} $\rrpp{}$ and \emph{with} $\rrpp{}$ on SIDD validation dataset.}
        (b) Even with the smallest inference stride factor ($b = 2$), BSN leaves unpleasing artifacts on the denoised results and cannot preserve the image structures well.
        (c) The proposed $\rrpp$ removes artifacts from BSN and significantly improves the denoising performances.
        For quantitative comparison, we also provide per-sample PSNR/SSIM \wrt ground-truth images at the bottom left of each patch.
    }
    \label{fig:rrpp_qualitative}
    \vspace{-4mm}
\end{figure*}